\documentclass[10pt,journal,compsoc]{IEEEtran}
\ifCLASSOPTIONcompsoc
  \usepackage[nocompress]{cite}
\else
  \usepackage{cite}
\fi
\ifCLASSINFOpdf
\else
\fi
\usepackage[utf8]{inputenc} 
\usepackage[T1]{fontenc}    
\usepackage{hyperref}       
\usepackage{url}            
\usepackage{amsfonts}       
\usepackage{nicefrac}       
\usepackage{microtype}      
\usepackage{graphicx}
\usepackage{mathtools}
\usepackage {multirow}
\usepackage{subfigure}
\usepackage{tabularx}
\usepackage{amsmath,amsfonts,amssymb}
\usepackage{wrapfig,lipsum,booktabs}
\usepackage{xcolor}
\usepackage{hyperref}
\newcommand{\vect}[1]{\mathbf{#1}}

\newcommand{\tabincell}[2]{\begin{tabular}{@{}#1@{}}#2\end{tabular}}

\def\eg{\textit{e.g.}}
\def\ie{\textit{i.e.}}

\def\etal{\textit{et al.}}

\newcommand{\amid}[1]{\textcolor{black}{#1}}

\usepackage{tikz}

\usepackage{enumitem}

\begin{document}
\title{Probabilistic Graph Attention Network with Conditional Kernels for Pixel-Wise Prediction}
%
%
%
%

\author{Dan Xu,
		Xavier Alameda-Pineda,~\IEEEmembership{Senior Member,~IEEE,}
        Wanli Ouyang,~\IEEEmembership{Senior Member,~IEEE,}
        Elisa Ricci,~\IEEEmembership{Member,~IEEE,}
        Xiaogang Wang,~\IEEEmembership{Senior Member,~IEEE,}
        Nicu Sebe,~\IEEEmembership{Senior Member,~IEEE}
\IEEEcompsocitemizethanks{
\IEEEcompsocthanksitem Dan Xu is with the Department of Computer Science and Engineering, Hong Kong University of Science and Technology. E-mail: danxu@cse.ust.hk.\protect
\IEEEcompsocthanksitem Xavier Alameda-Pineda is with the Perception Group, INRIA. E-mail: xavier.alameda-pineda@inria.fr.\protect
\IEEEcompsocthanksitem Wanli Ouyang is with the Department of Electrical and Information Engineering, the University of Sydney. E-mail: wanli.ouyang@sydney.edu.au.\protect
\IEEEcompsocthanksitem Xiaogang Wang is with the Department of Electronic Engineering, the Chinese University of Hong Kong. E-mail: xgwang@ee.cuhk.edu.hk
\IEEEcompsocthanksitem Elisa Ricci and Nicu Sebe are with the Department of Information Engineering and Computer Science, University of Trento, Italy. E-mail: \{elisa.ricci, niculae.sebe\}@unitn.it.\protect
}
\thanks{Manuscript received April 19, 2005; revised August 26, 2015.}}

%
%

\markboth{Journal of \LaTeX\ Class Files,~Vol.~14, No.~8, August~2015}%
{Shell \MakeLowercase{\textit{et al.}}: Bare Demo of IEEEtran.cls for Computer Society Journals}
%



\IEEEtitleabstractindextext{%
\begin{abstract}
Multi-scale representations deeply learned via convolutional neural networks have shown tremendous importance for various pixel-level prediction problems. In this paper we present a novel approach that advances the state of the art on pixel-level prediction in a fundamental aspect, \ie~structured multi-scale features learning and fusion. In contrast to previous works directly considering multi-scale feature maps obtained from the inner layers of a primary CNN architecture, and simply fusing the features with weighted averaging or concatenation, we propose a probabilistic graph attention network structure based on a novel Attention-Gated Conditional Random Fields (AG-CRFs) model for learning and fusing multi-scale representations in a principled manner. In order to further improve the learning capacity of the network structure, we propose to exploit feature dependant conditional kernels within the deep probabilistic framework. 
Extensive experiments are conducted on four publicly available datasets (\ie~BSDS500, NYUD-V2, KITTI and Pascal-Context) and on three challenging pixel-wise prediction problems involving both discrete and continuous labels (\ie~monocular depth estimation, object contour prediction and semantic segmentation). Quantitative and qualitative results demonstrate the effectiveness of the proposed latent AG-CRF model and the overall probabilistic graph attention network with feature conditional kernels for structured feature learning and pixel-wise prediction.
\end{abstract}
}

\maketitle

\IEEEdisplaynontitleabstractindextext

%
\IEEEpeerreviewmaketitle

\IEEEraisesectionheading{\section{Introduction}\label{sec:introduction}}

\begin{figure*}[!t]
\centering
\includegraphics[width=0.95\linewidth]{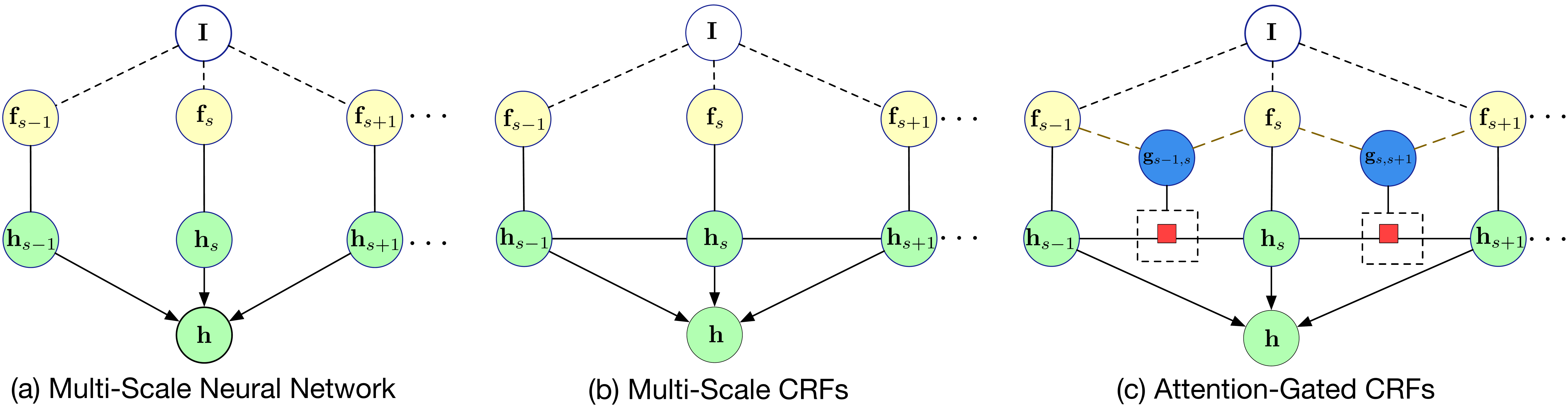}
\vspace{-8pt}
\caption{An illustration of different schemes for multi-scale deep feature learning and fusion. (a) the traditional approach (\textit{e.g.}\ concatenation, weighted averaging), (b) the proposed CRF implementing multi-scale feature fusion (c) the proposed Attention-Gated-CRF based approach.}
\vspace{-5pt}
\label{fig:AGMCL}
\end{figure*}

%
%
%
%
\IEEEPARstart{T}{he} 
capability to effectively 
exploit multi-scale feature representations is considered a crucial factor for achieving accurate predictions for the pixel-level prediction in both traditional~\cite{ren2008multi} and CNN-based~\cite{xie2015holistically,kokkinos2015pushing,maninis2016convolutional,chen2016deeplab} approaches.  
Restricting the attention on deep learning-based solutions, existing methods~\cite{xie2015holistically,maninis2016convolutional} typically
derive multi-scale representations by adopting standard CNN architectures and directly considering the feature maps associated to different inner semantic layers. 
These maps are highly complementary: while the features from the shallow layers are responsible for predicting low-level details, the ones from the deeper layers are devoted to encode the high-level semantic structure of the objects. Traditionally, concatenation and weighted average are very popular strategies to combine multi-scale representations (see Figure~\ref{fig:AGMCL}.a). While these strategies generally lead to an increased 
prediction accuracy with a comparison to single-scale models, they severely simplify the complex structured relationship between multi-scale feature maps. The motivational cornerstone of this study is the following research question: is it worth modelling and exploiting complex relationships between 
multiple scales of deep representations for pixel-wise prediction?

\begin{figure}[!t]
\centering
\includegraphics[width=0.9\linewidth]{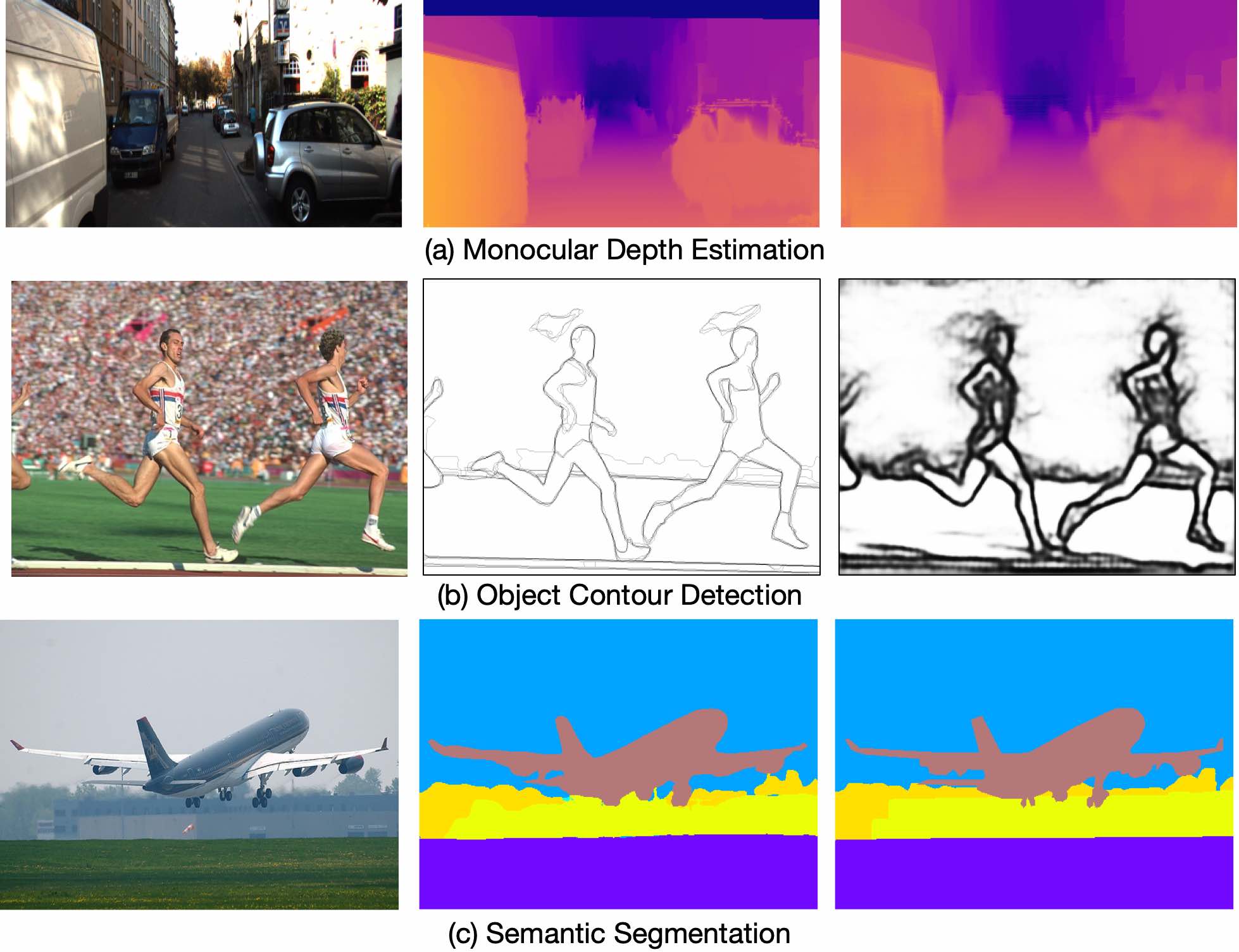}
\vspace{-10pt}
\caption{The proposed model targets multi-scale structured deep representation learning and could be applied into different pixel-wise prediction problems involving both discrete and continuous prediction variables, \ie~(a) monocular depth estimation on KITTI, (b) object contour detection on BSDS500, and (c) semantic segmentation on Pascal-Context. The first, second and third columns are input RGB images, ground-truth and predicted results, respectively.}
\vspace{-12pt}
\label{fig:motivation}
\end{figure}

Inspired by the success of recent works employing graphical models within deep CNN architectures~\cite{liu2015deep,xu2017multi} for structured prediction,  
we propose a probabilistic graph attention network structure base on a novel Attention-Gated Conditional Random Fields (AG-CRFs), which allows to learn effective feature representations at each scale by exploiting the information available from other scales. This is achieved by incorporating an attention mechanism~\cite{mnih2014recurrent} seamlessly integrated into the multi-scale learning process under the form of gates~\cite{minka2009gates}. Intuitively, the attention mechanism will further enhance the learning of the multi-scale representation fusion by controlling the information flow (\ie~messages) among the feature maps, thus improving the overall
performance of the model. To further increase the learning capacity of the model, we introduce feature dependant conditional kernels for predicting the attentions and the messages, enabling them conditioned on related feature context while not shared by all the feature inputs. \amid{In contrast to previous works~\cite{liu2015deep,xu2017multi} aiming at \amid{structured} modelling on the prediction level, our model focuses on the feature level, which leads to a much higher flexibility when applied to different pixel-wise prediction problems involving both continuous and discrete prediction variables.}

We implement the proposed AG-CRF as a neural network module, and integrate it into a hierarchical multi-scale CNN framework, defining a novel probabilistic graph attention network structure, termed as PGA-Net for pixel-wise prediction. The hierarchical network is able to learn richer multi-scale features than conventional CNNs, 
the representational power of which is further enhanced by the proposed conditional kernel AG-CRF model. We extensively evaluate the effectiveness of the proposed model on three different continuous and discrete pixel-wise prediction tasks (see Figure~\ref{fig:motivation}), \ie~object contour prediction, monocular depth estimation and semantic segmentation, and on multiple challenging benchmarks (BSDS500~\cite{arbelaez2011contour}, NYUD-V2~\cite{silberman2012indoor}, KITTI~\cite{Geiger2013IJRR} and Pascal-Context~\cite{mottaghi2014role}).  
The results demonstrate that our approach is able to learn rich and effective deep structured representations, thus showing very competitive performance to state-of-the-art methods on these tasks.

This paper extends our earlier work~\cite{xu2017learningdeep} through proposing 
a new feature dependant conditional kernel strategy, further re-elaborating the related works, providing more methodological details, and significantly expanding the experiments and analysis by demonstrating the effectiveness on another two popular pixel-wise prediction tasks (monocular depth estimation and semantic segmentation).   
Multi-scale deep features are widely demonstrated very effective. (\eg~\cite{hariharan2015hypercolumns,xie2015holistically}). The importance of our work is in joint probabilistic modelling of the relationship among multi-scale deep features using the conditional random fields and the attention mechanism. To summarize, the contribution of this paper is threefold:
\begin{itemize}[leftmargin=*]
\item First, we propose a structured probabilistic graph network for effectively learning and fusing multi-scale deep representations. We learn the multi-scale features using a probabilistic graphical attention model, which is a principled way of modelling the statistical relationship among multi-scale features.

\item Second, we design an attention guided CRF graph model which models the attention as gating for controlling the message passing among features of different scales. As the passed message among the feature maps are not always useful, the attention mechanism is especially introduced to control the message passing flow among feature maps of different scales. The attention is incorporated in the probabilistic graphical model. We also introduce a conditional kernel strategy for feature dependant attention and message learning.

\item Third, extensive experiments are conducted on three distinct pixel-wise prediction tasks and on four different challenging datasets, demonstrating that  the proposed
model and framework significantly outperform previous
methods integrating multi-scale information from different
semantic network layers, and show very competitive performance
on all the tasks compared with the state-of-the-art methods. The proposed model is generic in multi-scale feature learning and can be flexibly employed in other continuous and discrete prediction problems.
\end{itemize}

\par The remainder of this paper is organized as follows. Sec.~\ref{sec:rel} introduces the related works, and then we illustrate the proposed approach in Sec.~\ref{sec:method}, and in Sec.~\ref{sec:networkimplementation} we present the details of the model implementation in deep networks. The experimental evaluation and analysis are elaborated in~Sec.~\ref{sec:experiments}. We finally conclude the paper in Sec.~\ref{sec:conclusion}.

\section{Related Work}\label{sec:rel} 
\subsection{Pixel-wise Prediction} We review previous works with deep learning networks on three important pixel-wise prediction tasks, \ie~contour detection, monocular depth estimation and semantic segmentation, on which we extensively demonstrate the effectiveness of the proposed approach.  
\par\noindent\textbf{Contour Detection.} 
In the last few years several deep learning models have been proposed for detecting 
contours~\cite{shen2015deepcontour,bertasius2015deepedge,yang2016object,xie2015holistically,maninis2016convolutional,liu2016richer} or \amid{crisp boundaries~\cite{wang2018deep}}.
Among these, some works explicitly focused on devising multi-scale CNN models in order to boost performance. For instance, the Holistically-Nested Edge Detection 
method~\cite{xie2015holistically} employed multiple side outputs derived from the inner layers of a primary CNN and combine them for the final prediction.
Liu~\etal~\cite{liu2016richer} introduced a framework to learn rich deep representations by concatenating features derived from all convolutional 
layers of VGG16. Bertasius \etal~\cite{bertasius2015deepedge} considered skip-layer CNNs to jointly combine feature maps from multiple layers. Maninis 
\etal~\cite{maninis2016convolutional} 
proposed Convolutional Oriented Boundaries (COB), where features from different layers are fused to compute oriented contours and region hierarchies. 
{However, these works combine the multi-scale representations from different layers adopting concatenation and weighted averaging schemes while not considering the dependency between the features. 
} 

\noindent\textbf{Monocular Depth Estimation.} 
There are existing recent works on monocular depth estimation based on {deep CNNs~\cite{eigen2015predicting,liu2015deep,wang2015towards,roymonocular,laina2016deeper, fu2018deep, gan2018monocular,lee2019big}}.
Among them, Eigen~\etal~\cite{eigen2014depth} proposed a multi-scale network architecture for the task via considering two cascaded networks, performing a coarse to fine refinement of the depth prediction. They also further extend this framework to deal with multiple pixel-level predictions, such as surface normal estimation and semantic segmentation. \amid{Fu~\etal~\cite{fu2018deep} presented a novel DORN method to cast the monocular depth estimation as a deep ordinal regression problem. 
Lee~\etal~\cite{lee2019big} designed a network module which utilizes the multi-scale local planar as guidance to learn more effective structure features for depth estimation.}
Wang~\etal~\cite{wang2015towards} introduced a CNN-CRF framework for joint depth estimation and semantic segmentation. The most related work to ours is~\cite{liu2015deep}, which introduced a continuous CRF model for end-to-end learning depth regression with a front-end CNN. Xu~\etal~\cite{xu2017multi} improved~\cite{liu2015deep} by presenting a multi-scale continuous CRF model to learn the multi-scale predictions and fusion. 
\amid{However, these approaches purely focus on modelling the structure of the predictions, therefore answering to specific problems, and leading to task-dependent models and associated architectures. Differently, our work is focusing on statistical modeling on the structured features, thus being more flexible to be applied to different continuous or discrete prediction tasks.}


\par\noindent\textbf{Semantic Segmentation.} 
As an important task in scene understanding, semantic segmentation has received wide attention in recent years. Long~\etal~\cite{long2015fully} proposed a fully convolutional network for the task which significantly improved the performance and reduced the network parameters. The dilated convolution~\cite{chen2016deeplab, yu2015multi} was devised in order to obtain bigger receptive field of the features, further boosting the segmentation performance. \amid{OCNet~\cite{yuan2018ocnet} introduced an object-context pooling strategy based on affinity learning among pixels to capture a global context for feature refinement.} Other main-stream directions mainly explored multi-scale feature learning and model ensembling~\cite{chen2016attention, hariharan2015hypercolumns}, designing convolutional encoder-decoder network structures~\cite{noh2015learning, badrinarayanan2015segnet} and performing end-to-end structure prediction with CRF models~\cite{liu2015semantic, arnab2016higher, zheng2015conditional}. A more close work to ours in the literature is the GloRe approach~\cite{chen2019graph} which utilizes a graph convolution model for learning generic representation for semantic segmentation. However, its modelling is only for single-scale and not in a probabilistic graph formulation.

\amid{There are also some existing works which explored joint deep learning of more than one pixel-wise prediction tasks~\cite{xu2018pad,zhang2018joint,zhang2019pattern,vandenhende2020mti}. Specifically, Xu~\etal~\cite{xu2018pad} designed a PAD-Net architecture which learns multiple auxiliary pixel-wise tasks and presents a multi-task distillation module to combine the predictions from the different auxiliary tasks to help more important final tasks. Vandenhende and Van Gool~\etal~\cite{vandenhende2020mti} further improved the PAD-Net by introducing a method for feature propagation of different multi-task representations. However, these works are focusing more on empirical design for learning interaction between different pixel-wise tasks, while our model targets at statistical probabilistic modeling for structured multi-scale feature fusion which could provide a theoretic explanation and thus it is beneficial for a more effective deep module design.}
\subsection{Deep Multi-scale Learning} 
The importance of combining multi-scale information has been widely revealed in various computer vision tasks~\cite{xie2015holistically}
For instance, Xie~\etal~\cite{xie2015holistically} proposed a fully convolutional neural network with deep supervision for edge detection, which employs a weighted averaging strategy for the combination of multi-scale side outputs. 
The skipping-layer networks are also very popular for learning multi-scale representations, where the features obtained from different semantic layers of a backbone CNN are combined in an output layer to produce more robust representations.
\amid{Sun and Wang~\etal~\cite{sun2019high, wang2020deep} proposed a HRNet architecture which aims to enhance high-resolution representations via aggregating the representations with multi-scale resolutions from different network stages.}
To aggregate multi-scale contexts of features, the \textit{dilation} or \textit{\`{a} trous} convolution~\cite{chen2014semantic} structures are devised, which could be applied in embedded in different convolutional layers in a deep network to obtain a larger receptive field.  
Yang~\etal~\cite{yang2015multi} introduced DAG-CNNs to combine multi-scale features produced from different ReLU layers using element-wise addition operation.
Huang~\etal~\cite{huang2018multi} recently proposed a multi-scale network architecture using densely skipping connections to pass feature flow at different scales. 
However, in these works, the multi-scale representations or predictions are typically combined via using simple concatenation or weighted averaging operation. 
We are also not aware of previous works exploring multi-scale representation learning and fusion within a probabilistic CRF graph framework. Besides, we also involve learning an attention mechanism as gating in the graph model for controlling the message passing of the feature variables.
 
\subsection{Attention Models}
Attention models~\cite{velivckovic2017graph} have been successfully exploited in deep learning for various tasks such 
as image classification~\cite{xiao2015application}, speech recognition~\cite{chorowski2015attention}, image caption generation~\cite{xu2015show} and language translation~\cite{vaswani2017attention}. \amid{Fu~\etal~\cite{fu2019dual} recently proposed a dual attention model considering a combination of both the spatial- and the channel-wise attentions for semantic segmentation.} However, 
to our knowledge, this work is the first to introduce a structured attention model for \amid{for both discrete and continuous prediction tasks}. Furthermore, we are not aware of previous studies integrating the attention mechanism into a probabilistic (CRF) framework to control the message passing between hidden variables. We model the attention as \textit{gates}~\cite{minka2009gates}, which have been used in previous deep models such as
restricted Boltzman machine for unsupervised feature learning~\cite{tang2010gated}, LSTM for sequence 
learning~\cite{chung2014empirical} and 
CNN for image classification~\cite{zeng2016crafting}. However, none of these works explores the possibility of jointly learning multi-scale deep representations
and an attention model within a unified probabilistic graphical model.
\amid{\subsection{Structured Learning based on CRFs}
The conditional random fields (CRFs) were widely used for probabilistic structured modeling in the non-deep-learning era for a wide range of problems, such as object recognition~\cite{quattoni2005conditional}, information extraction~\cite{sarawagi2005semi} and pixel-wise semantic labeling~\cite{lafferty2001conditional, he2004multiscale, krahenbuhl2011efficient, boykov2006graph}. Specifically, Boykov~\etal~\cite{boykov2006graph} proposed a combinational graph cut algorithm integrating cues of boundaries, regions and shapes for semantic segmentation. Kr{\"a}henb{\"u}hl and Koltun~\cite{krahenbuhl2011efficient} designed a fully-connected CRF model with gaussian pair-wise potentials and accordingly proposed an efficient approximation inference solution for the model. With the rapid progress of the deep learning techniques, the CRFs are also utilized together with Convolutional Neural Network (CNN) architectures for learning structured deep predictions~\cite{zheng2015conditional, liu2015deep,arnab2016higher}. Among them, Zheng~\etal~\cite{zheng2015conditional} first implemented the CRF inference as Recurrent Neural Networks for end-to-end learning with any backbone CNN. There are also existing works exploring learning structured deep features with CRFs, in order to more flexibly adapt to different continuous and discrete tasks, such as human pose estimation~\cite{chu2016crf} and monocular depth estimation~\cite{liu2015deep}. Great success has been made by these existing models; however, none of them considered simultaneously learning structured multi-scale representations and structured attention in a joint probabilistic graph formulation for pixel-wise prediction.} 


\section{The Proposed Approach}\label{sec:method}
\subsection{Problem Definition and Notation}
Given an input image $\vect{I}$ and a generic front-end CNN model with parameters $\vect{W}_c$, we consider a set of 
$S$ multi-scale feature maps $\vect{F} = \{\vect{f}_{s}\}_{s=1}^{S}$. Being a generic framework, these feature maps can 
be the output of $S$ intermediate CNN layers or of another representation, thus $s$ is a \textit{virtual} scale. 
The feature map at 
scale $s$, $\vect{f}_{s}$ can be interpreted as a set of feature vectors, $\vect{f}_{s} = 
\{\vect{f}_{s}^i\}_{i=1}^{N}$, where $N$ is the number of pixels. Opposite to previous works adopting simple 
concatenation or weighted averaging schemes 
\cite{hariharan2015hypercolumns, xie2015holistically}, we propose to combine the multi-scale feature maps by learning a 
set of latent feature maps $\vect{h}_{s} = \{\vect{h}_{s}^i\}_{i=1}^{N}$ with a novel \textit{Attention-Gated CRF} 
model sketched in Figure\ref{fig:AGMCL}. Intuitively, this allows a joint refinement of the features by flowing 
information between different scales. Moreover, since the information from one scale 
may or may not be relevant for the pixels at another scale, we utilise the concept of \textit{gate}, previously 
introduced in the literature in the case of graphical models~\cite{winn2012causality}, in our CRF formulation. These 
gates are binary random hidden variables that permit or block the flow of information between scales at every pixel. 
Formally, $g_{s_e,s_r}^i\in\{0,1\}$ is the gate at pixel $i$ of scale $s_r$ (receiver) from scale $s_e$ (emitter), and 
we also write $\vect{g}_{s_e,s_r} = \{g_{s_e,s_r}^i\}_{i=1}^N$. Precisely, when $g_{s_e,s_r}^i=1$ then the hidden 
variable $\vect{h}_{s_r}^i$ is updated taking (also) into account the information from the $s_e$-th layer, 
\textit{i.e.}\ 
$\vect{h}_{s_e}$. As shown in the following, the joint inference of the hidden features and the gates leads 
to estimating the optimal features as well as the corresponding attention model, hence the name Attention-Gated CRFs.

\begin{figure*}[!t]
\centering
\includegraphics[width=0.9\linewidth]{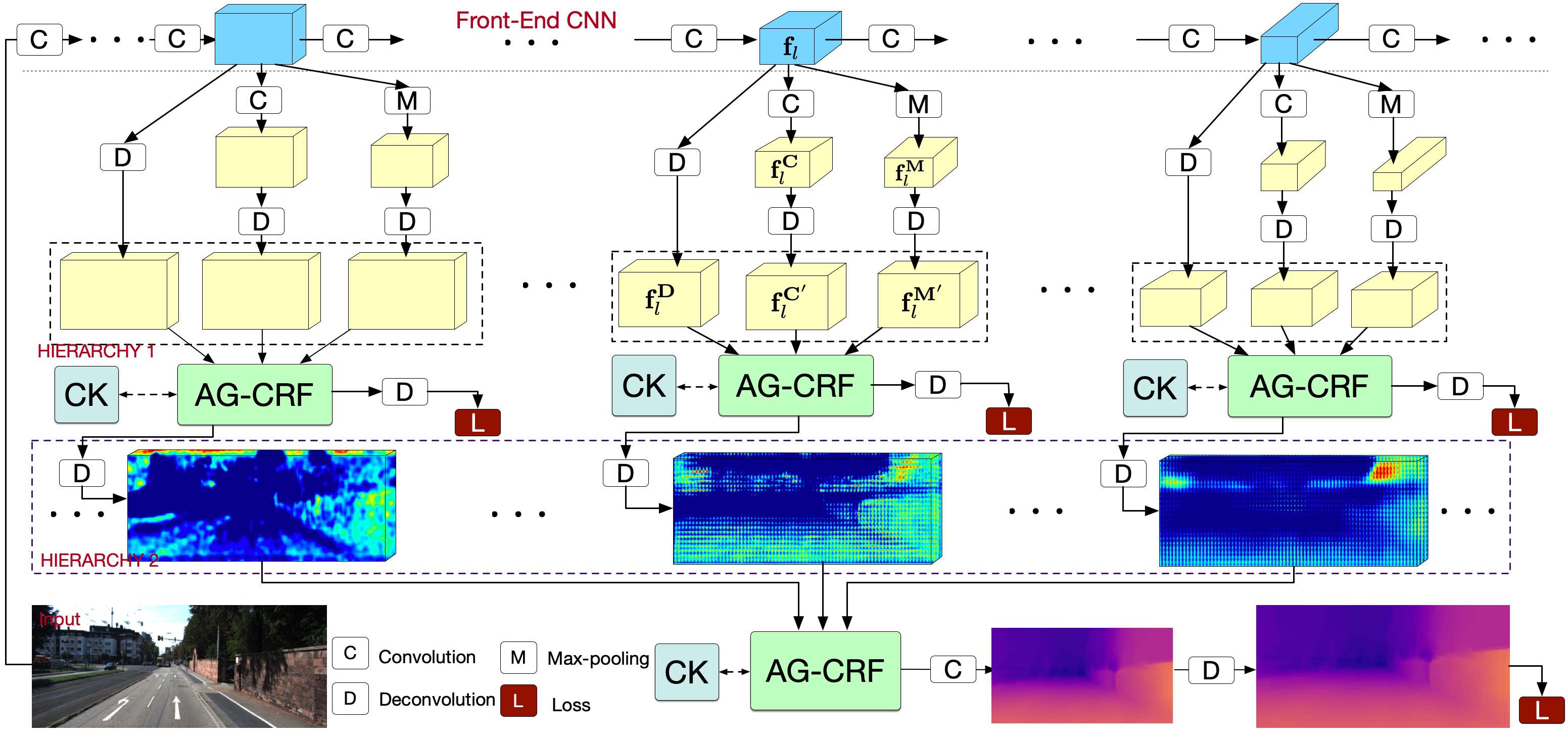}
\caption{An overview of the profposed Probabilistic Graph Attention Network (PGA-Net) for monocular depth detection. The symbols $\mathrm{C}$, $\mathrm{D}$, $\mathrm{M}$ and $\mathrm{L}$ denote the convolution, the deconvolution, the max-pooling operation and optimization loss, respectively. AG-CRF represents the proposed attention-gated CRF model with conditional kernels (CK) for structured multi-scale feature learning, which is fully differentiable and supports end-to-end training with a multi-scale CNN network. PGA-Net consists of two hierarchies. The hierarchy 1 generates rich multi-scale features which are refined by AG-CRFs, and then are passed to hierarchy 2 for final prediction. }
\label{framework}
\end{figure*}

\subsection{Attention-Gated CRFs}
\label{sec:attention}
Given the observed multi-scale feature maps $\vect{F}$ of image $\vect{I}$, the objective is to estimate the hidden 
multi-scale representation $\vect{H}=\{\vect{h}_{s}\}_{s=1}^S$ and, accessorily the attention gate variables 
$\vect{G}=\{\vect{g}_{s_e,s_r}\}_{s_e,s_r=1}^{S}$. To do that, we formalize the problem within a conditional random 
field framework and write the Gibbs distribution as %
\begin{equation}
P(\vect{H}, \vect{G} |  \vect{I}, \Theta) = 
 \mathrm{exp}\left(-E(\vect{H}, \vect{G}, \vect{I}, \Theta)\right)/Z\left(\vect{I}, \Theta \right),
  \end{equation}
where $\Theta$ is the set of parameters and $E$ is the energy function. As usual, we exploit both unary and binary 
potentials to couple the hidden variables between them and to the observations. Importantly, the 
proposed binary potential is gated, and thus only active when the gate is open. More formally the general 
form\footnote{One could certainly include a unary potential for the gate variables as well. However this would imply 
that there is a way to set/learn the a priori distribution of opening/closing a gate. In practice we did not observe 
any notable difference between using or skipping the unary potential on $g$.} of the energy function writes:
\begin{equation}
\begin{aligned}
 E&(\vect{H}, \vect{G}, \vect{I}, \Theta) = \\
 &\underbrace{\sum_{s} \sum_{i} \phi_h (\vect{h}_{s}^i, \vect{f}_{s}^i)}_{\rm{Unary~potential}}
+ \underbrace{ \sum_{s_e,s_r} \sum_{i,j} g_{s_e,s_r}^i \psi_h (\vect{h}_{s_r}^i, \vect{h}_{s_e}^j,{\mathbf{K}_{s_e,s_r}^{i,j}}) 
}_{\rm{Gated~pairwise~potential}}\\
\end{aligned}
\label{equ:ag-crf}
\end{equation}

The first term of the energy function is a classical unary term that relates the hidden features to the observed 
multi-scale CNN representations. The second term synthesizes the theoretical contribution of the present study because 
it conditions the effect of the pair-wise potential $\psi_h (\vect{h}_{s_e}^i, \vect{h}_{s_r}^j)$ upon the gate hidden 
variable $g_{s_e,s_r}^i$. Figure~\ref{fig:AGMCL}c depicts the model formulated in Equ.(\ref{equ:ag-crf}). If we remove the attention gate variables, it becomes a general multi-scale CRFs as shown in Figure~\ref{fig:AGMCL}b.
\par Given that formulation, and as it is typically the case in conditional random fields, we 
exploit the mean-field approximation in order to derive a tractable inference procedure. Under this generic form, the 
mean-field inference procedure writes:
\begin{equation}
\begin{aligned}
q(\vect{h}_{s}^i) \propto &\exp\Big(\phi_h(\vect{h}_s^i,\vect{f}_s^i) + \\ & \sum_{s'\neq s}\sum_{j} 
\mathbb{E}_{q(g_{s',s}^i)}\{g_{s',s}^i\}  \mathbb{E}_{q(\vect{h}_{s'}^j)}\{\psi_h(\vect{h}_{s}^i, \vect{h}_{s'}^j)\} 
\Big),
\end{aligned}
\end{equation}
\begin{equation}
q(g_{s',s}^i) \propto \exp\Big( g_{s',s}^i \mathbb{E}_{q(\vect{h}_{s}^i)}\Big\{ \sum_{j} 
\mathbb{E}_{q(\vect{h}_{s'}^j)} \left\{\psi_h(\vect{h}_{s}^i, \vect{h}_{s'}^j)\right\} \Big\} \Big),
\end{equation}
where $\mathbb{E}_q$ stands for the expectation with respect to the distribution $q$.

Before deriving these formulae for our precise choice of potentials, we remark that, since the gate is a binary 
variable, the expectation of its value is the same as $q(g_{s',s}^i=1)$. By defining: ${\cal M}_{s',s}^i = 
\mathbb{E}_{q(\vect{h}_{s}^i)}\left\{ \sum_{j} 
\mathbb{E}_{q(\vect{h}_{s'}^j)} \left\{\psi_h(\vect{h}_{s}^i, \vect{h}_{s'}^j)\right\} \right\}$, the expected value of 
the gate writes:
\begin{equation}
\begin{aligned}
\alpha_{s,s'}^i &= \mathbb{E}_{q(g_{s',s}^i)}\{g_{s',s}^i\} = \frac{q(g_{s',s}^i=1)}{q(g_{s',s}^i=0)+q(g_{s',s}^i=1)} \\ &= 
\sigma\left(-{\cal M}_{s',s}^i\right),
\end{aligned}
\end{equation}
where $\sigma()$ denotes the sigmoid function. This finding is specially relevant in the framework of CNN since many of 
the attention models are typically obtained after applying the sigmoid function to the features derived from a 
feed-forward network. Importantly, since the quantity ${\cal M}_{s',s}^i$ depends on the expected values of the hidden 
features $\vect{h}_{s}^i$, the AG-CRF framework extends the unidirectional connection from the features to the attention 
model, to a bidirectional connection in which the expected value of the gate allows to refine the distribution of the 
hidden features as well.

\subsection{AG-CRF Inference}
In order to construct an operative model we need to define the unary and gated potentials $\phi_h$ and 
$\psi_h$. In our case, the unary potential corresponds to an isotropic Gaussian:
\begin{equation}
 \phi_h(\vect{h}_{s}^i,\vect{f}_{s}^i) = - \frac{a_s^i}{2}\|\vect{h}_{s}^i-\vect{f}_s^i\|^2,
\end{equation}
where $a_s^i>0$ is a weighting factor.

The gated binary potential is specifically designed for a two-fold objective. On the one hand, we would like to learn 
and further exploit the relationships between hidden vectors at the same, as well as at different scales. On the other 
hand, we would like to exploit previous knowledge on attention models and include linear terms in the potential. 
Indeed, this would implicitly shape the gate variable to include a linear operator on the features. Therefore, we chose a bilinear potential:
\begin{equation}
 \psi_h(\vect{h}_s^i,\vect{h}_{s'}^j) = \tilde{\vect{h}}_s^i \vect{K}_{s,s'}^{i,j} \tilde{\vect{h}}_{s'}^j,
\end{equation}
where $\tilde{\vect{h}}_s^i = (\vect{h}_s^{i \top}, 1)^\top$ and $\vect{K}_{s,s'}^{i,j}\in\mathbb{R}^{ (C_s+1) \times 
(C_{s'}+1)}$ being $C_s$ the size, i.e. the number of channels, of the representation at scale $s$. If we write this 
matrix as $\vect{K}_{s,s'}^{i,j} = ( \vect{L}_{s,s'}^{i,j}, \vect{l}_{s,s'}^{i,j} ; \vect{l}_{s',s}^{j,i \top}, 1 )$, 
then $\vect{L}_{s,s'}^{i,j}$ exploits the relationships between hidden variables, while $\vect{l}_{s,s'}^{i,j}$ and 
$\vect{l}_{s',s}^{j,i}$ implement the classically used linear relationships of the attention models. In order words, $\psi_h$ models the 
pair-wise relationships between features with the upper-left block of the matrix. 
Furthermore, $\psi_h$ takes into account the linear relationships by completing the hidden vectors with the unity. In 
all, the energy function writes:
\begin{equation}
\begin{aligned}
 E&(\vect{H}, \vect{G}, \vect{I}, \Theta) =  \\ 
 &- \sum_{s} \sum_{i} \frac{a_s^i}{2} \|\vect{h}_{s}^i-\vect{f}_s^i\|^2 + 
 \sum_{s_e,s_r} \sum_{i,j} g_{s_e,s_r}^i \tilde{\vect{h}}_{s_r}^i \vect{K}_{s_r,s_e}^{i,j} \tilde{\vect{h}}_{s_e}^j.
 \end{aligned}
\end{equation}
Under these potentials, we can consequently update the mean-field inference equations to:
\begin{equation}
\begin{aligned}
 q(\vect{h}_s^i) \propto &\exp\Big( -\frac{a_s^i}{2}(\|\vect{h}_s^i\| - 2\vect{h}_s^{i \top}\vect{f}_s^i) + \\ 
& \sum_{s'\neq s} \alpha_{s,s'}^i \vect{h}_s^{i \top}\sum_{j} (\vect{L}_{s,s'}^{i,j} \bar{\vect{h}}_{s'}^j + \vect{l}_{s,s'}^{i,j})
%
\Big),
\end{aligned}
\end{equation}
where $\bar{\vect{h}}_{s'}^j$ is the expected a posterior value of $\vect{h}_{s'}^j$. 

The previous expression implies that the a posterior distribution for $\vect{h}_s^i$ is a Gaussian. The mean vector 
of the Gaussian and the function ${\cal M}$ write:
\begin{equation}\label{eq:inferenceFL}
\begin{aligned}
& \bar{\vect{h}}_{s}^i = \frac{1}{a_s^i}\Big(a_s^i\vect{f}_s^i + \sum_{s'\neq s}\alpha_{s,s'}^i \sum_{j} 
(\vect{L}_{s,s'}^{i,j} \bar{\vect{h}}_{s'}^j + \vect{l}_{s,s'}^{i,j}) \Big) \quad \\
&  {\cal M}_{s',s}^i = \sum_{j} \left( \bar{\vect{h}}_s^i \vect{L}_{s,s'}^{i,j} \bar{\vect{h}}_{s'}^j + 
\bar{\vect{h}}_s^{i \top}\vect{l}_{s,s'}^{i,j} + \bar{\vect{h}}_{s'}^{j \top} \vect{l}_{s',s}^{j,i} \right)
\end{aligned}
\end{equation}
which concludes the inference procedure. 
Furthermore, the proposed framework can be simplified to obtain the traditional attention models. In most 
of the previous studies, the attention variables are computed directly from the multi-scale features instead of 
computing them from the hidden variables. Indeed, since many of these studies do not propose a probabilistic 
formulation, there are no hidden variables and the attention is computed sequentially through the scales. We can 
emulate the same behavior within the AG-CRF framework by modifying the gated potential as follows:
\begin{equation}\label{eq:9}
 \tilde{\psi}_h(\vect{h}_s^i,\vect{h}_{s'}^j,\vect{f}_s^i,\vect{f}_{s'}^j) = \vect{h}_s^i \vect{L}_{s,s'}^{i,j} 
\vect{h}_{s'}^j + \vect{f}_s^{i \top}\vect{l}_{s,s'}^{i,j} + \vect{f}_{s'}^{j \top} \vect{l}_{s',s}^{j,i}.
\end{equation}
This means that we keep the pair-wise relationships between hidden variables (as in any CRF) and let the attention model 
be generated by a linear combination of the observed features from the CNN, as it is traditionally done. The changes in the inference procedure are straightforward.
We refer to this model as 
partially-latent AG-CRFs (PLAG-CRFs), whereas the more general one is denoted as fully-latent AG-CRFs (FLAG-CRFs).
The potential defined for the PLAG-CRF model, \ie~(\ref{eq:9}), has an impact on the inference of both the hidden features and the attention gates. Indeed, since the linear term does not depend on the hidden features, but on the observations, the mean of the hidden features is computed independently of this linear term:
\begin{equation}\label{eq:inferencePL1}
 \bar{\vect{h}}_{s}^i = \frac{1}{a_s^i}\Big(a_s^i\vect{f}_s^i + \sum_{s'\neq s}\alpha_{s,s'}^i \sum_{j} 
\vect{L}_{s,s'}^{i,j} \bar{\vect{h}}_{s'}^j \Big).
\end{equation}
Likewise, the linear terms of the attention gate do not depend anymore on the hidden features but on the observations from the CNN:
\begin{equation}\label{eq:inferencePL2}
 {\cal M}_{s',s}^i = \sum_{j} \left( \bar{\vect{h}}_s^i \vect{L}_{s,s'}^{i,j} \bar{\vect{h}}_{s'}^j + 
\bar{\vect{f}}_s^{i \top}\vect{l}_{s,s'}^{i,j} + \bar{\vect{f}}_{s'}^{j \top} \vect{l}_{s',s}^{j,i} \right).
\end{equation}
These are the two differences with respect to the inference of FLAG-CRFs. We will also introduce the implementation difference of both versions in Sec.~\ref{sec:networkimplementation}.

\subsection{Feature Dependant Conditional Kernels} In the model inference as shown in~(\ref{eq:inferenceFL}),~(\ref{eq:inferencePL1}) and~(\ref{eq:inferencePL2}), the kernels $\vect{L}_{s,s'}^{i,j}$, $\vect{l}_{s,s'}^{i,j}$ and $\vect{l}_{s',s}^{j,i}$ are shared for all the input features. This property restricts the learning capacity of the model: one would like those kernels to be dependant on the features so as to capture the feature correlated context, which is particularly important for pixel-wise prediction tasks. We hence propose to learn feature conditioned kernels, instead of the previous shared kernels. In practice, each kernel is predicted from the input features using a linear transformation as follows:
\begin{equation}\label{eq:featureconditionalkernel}
\begin{aligned}
&\vect{L}_{s,s'}^{i,j} = {\vect{W}_L}_{s,s'}^{i,j} \mathrm{concat}(\bar{\vect{h}}_s^i, \bar{\vect{h}}_{s'}^j) + {\vect{b}_L}_{s,s'}^{i,j},\\
&\vect{l}_{s,s'}^{i,j} = {\vect{W}_l}_{s,s'}^{i,j}\ \bar{\vect{h}}_s^i + {\vect{b}_l}_{s,s'}^{i,j}, \quad
\vect{l}_{s',s}^{i,j} = {\vect{W}_l}_{s',s}^{i,j} \bar{\vect{h}}_{s'}^j + {\vect{b}_l}_{s',s}^{i,j},
\end{aligned}
\end{equation}
where $\mathrm{concat(\cdot)}$ denotes a concatenation operation function. The symbols $\{\vect{{W}_L}_{s,s'}^{i,j}, {\vect{b}_L}_{s,s'}^{i,j}\}$, $\{{\vect{W}_l}_{s,s'}^{i,j}, {\vect{b}_l}_{s,s'}^{i,j}\}$ and $\{{\vect{W}_l}_{s',s}^{i,j}, {\vect{b}_l}_{s',s}^{i,j}\}$ are the parameters of the linear transformation. By making this concise modification in the model inference, we further clearly boost the performance of the model on different pixel-wise prediction tasks, which will be elaborated in the experimental part.

\section{Network Implementation}\label{sec:networkimplementation}
\subsection{Neural network implementation for joint learning}
\begin{figure}[!t]
\centering
\includegraphics[width=0.9\linewidth]{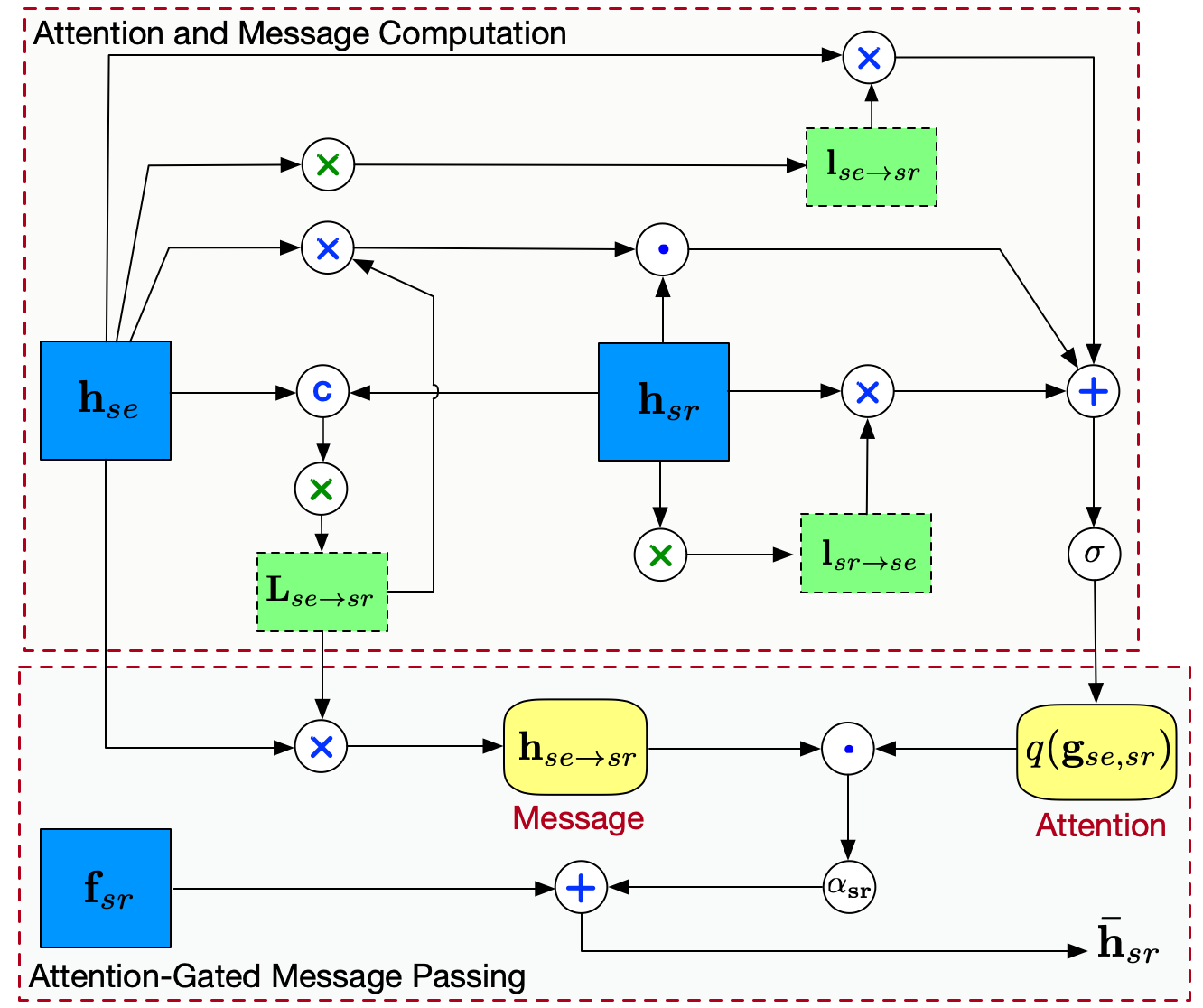}
\caption{The detailed computing flow of the mean-field updating of the proposed conditional kernel AG-CRF model. The symbol $\otimes$ denotes the convolutional operation. The ones with green color represent the operation for the conditional kernel prediction. The symbols $\odot$ and $\oplus$ denote element-wise multiplication and addition operation, respectively. The symbols \textcircled{$\sigma$} and \textcircled{c} represent a sigmoid and a concatenation operation, respectively.}
\label{fig:computingflow}
\vspace{-5pt}
\end{figure}

In order to infer the hidden variables and learn the parameters of the AG-CRFs together with those of the front-end CNN, we implement the AG-CRFs updates in neural network. A detailed computing flow is depicted in Figure~\ref{fig:computingflow}. The implementation consists of several steps: 
\begin{itemize}[leftmargin=*]
\item Conditional kernel prediction for the kernels $\vect{L}_{s_e \rightarrow s_r}$, $\vect{l}_{s_e \rightarrow s_r}$, and $\vect{l}_{s_r \rightarrow s_e}$ with $\vect{L}_{s_e \rightarrow s_r} \leftarrow {\vect{W}_L}_{s_e \rightarrow s_r} \otimes (\vect{h}_{s_e} \text{\textcircled{c}} \vect{h}_{s_r}) $, 
$\vect{l}_{s_e \rightarrow s_r} \leftarrow {\vect{W}_l}_{s_e \rightarrow s_r} \otimes \vect{h}_{s_e}$,
and 
$\vect{l}_{s_r \rightarrow s_e} \leftarrow {\vect{W}_l}_{s_r \rightarrow s_e} \otimes \vect{h}_{s_r}$;

\item Message passing from the $s_e$-th scale to the current $s_r$-th 
scale is performed with $\vect{h}_{s_e\rightarrow s_r} \leftarrow \vect{L}_{s_e \rightarrow s_r} 
\otimes \vect{h}_{s_e}$, where $\otimes$ denotes the convolutional operation and $\mathbf{L}_{s_e \rightarrow s_r}$ 
denotes the corresponding convolution kernel;

\item Attention map estimation $q(\vect{g}_{s_e, s_r}=\vect{1}) \leftarrow 
\sigma(\vect{h}_{s_r} \odot (\vect{L}_{s_e \rightarrow s_r} \otimes \vect{h}_{s_e} ) 
+ \mathbf{l}_{s_e \rightarrow s_r} \otimes \mathbf{h}_{s_e} + \mathbf{l}_{s_r\rightarrow s_e} \otimes 
\mathbf{h}_{s_r})$, where $\mathbf{L}_{s_e \rightarrow s_r}$, $\vect{l}_{s_e\rightarrow s_r}$ and 
$\vect{l}_{s_r\rightarrow s_e}$ are convolution kernels and $\odot$ represents element-wise product operation;
\item Attention-gated message 
passing from other scales and adding unary term: $\bar{\vect{h}}_{s_r} = \mathbf{f}_{s_r} \oplus a_{s_r} \sum_{s_e \neq s_r} ( q(\mathbf{g}_{s_e, 
s_r}=\vect{1}) \odot \vect{h}_{s_e\rightarrow s_r})$, where $a_{s_r}$ encodes the effect of the $a_{s_r}^i$ for weighting the message and can be 
implemented as a $1\times 1$ convolution. The symbol $\oplus$ denotes element-wise addition. 
\end{itemize}
In order to simplify the overall inference procedure, and because the magnitude of the linear term of 
$\psi_h$ is in practice negligible compared to the quadratic term, we discard the message associated to the linear term. When the inference is 
complete, the final estimate is obtained by convolving all the scales. For the inference of the partially latent model, we only need to discard the corresponding terms in the computation of the messages in the second step, and replace the latent features with observation features for the attention prediction in the third step.

\subsection{Exploiting AG-CRFs with a Multi-scale Network}
\textbf{PGA-Net Architecture.} The proposed Attention-guided Multi-scale Hierarchical Network (PGA-Net), as sketched in 
Figure~\ref{framework}, consists of a multi-scale hierarchical network (MH-Net) together with the AG-CRF model described 
above. The MH-Net can be constructed from a front-end CNN architecture such as the widely used 
AlexNet~\cite{krizhevsky2012imagenet}, VGG~\cite{simonyan2014very} and ResNet~\cite{He2015}. One prominent feature of 
MH-Net is its ability to generate richer multi-scale representations. In order to do that, we perform distinct non-linear 
mappings (deconvolution $\vect{D}$, convolution $\vect{C}$ and max-pooling $\vect{M}$) upon $\vect{f}_l$, the CNN feature 
representation from an intermediate layer $l$ of the front-end CNN. This leads to a three-way representation: 
$\vect{f}_{l}^\vect{D}$, $\vect{f}_{l}^\vect{C}$ and $\vect{f}_{l}^\vect{M}$. Remarkably, while $\vect{D}$ upsamples the feature map, $\vect{C}$ maintains its original size and $\vect{M}$ reduces it, and different kernel size is utilized for them to have different receptive fields, then naturally obtaining complementary inter- and multi-scale representations. The $\vect{f}_{l}^\vect{C}$ and $\vect{f}_{l}^\vect{M}$ are further aligned to the dimensions of the feature map $\vect{f}_{l}^\vect{D}$ by the deconvolutional operation. The hierarchy is implemented in two levels. The first level uses an AG-CRF model 
to fuse the three representations of each layer $l$, thus refining the CNN features within the same scale. The second level of the hierarchy uses 
an AG-CRF model to fuse the information coming from multiple CNN layers. The proposed hierarchical multi-scale structure is general purpose and 
able to involve an arbitrary number of layers and of diverse intra-layer representations. \amid{It should be also noted that the proposed AG-CRF model is flexible to be applied into any multi-scale context in a deep learning network for structured representation refinement.}

\begin{figure}[!t]
\centering
\includegraphics[width=0.95\linewidth]{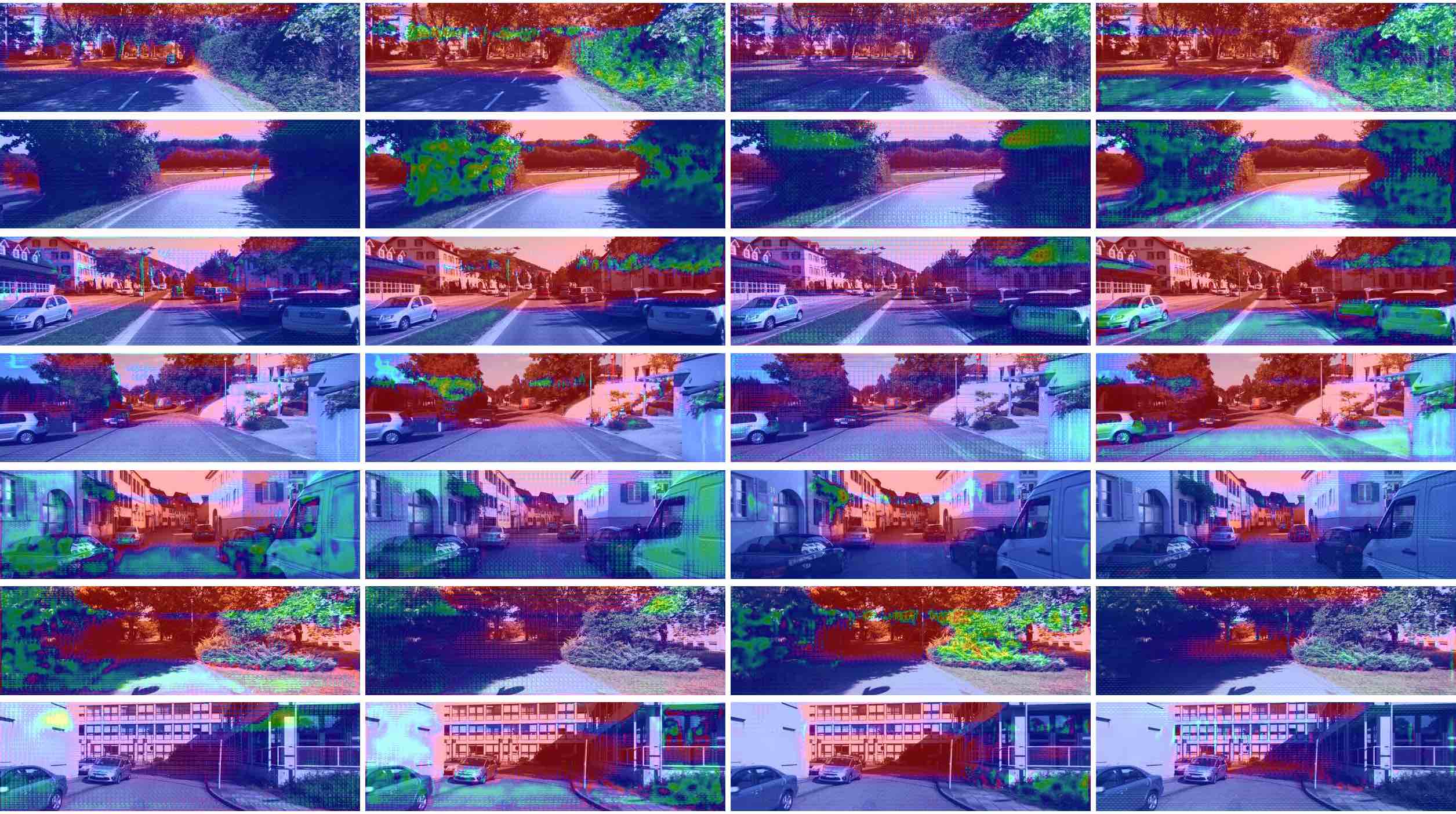}
\vspace{-5pt}
\caption{The visualization of the learned attention maps in the proposed AG-CRF model. Our attention is a pixel-wise attention, \ie~simultaneously learning both spatial- and channel-wise attention. We visualize the attention by uniformly sampling four attention channels of the attention map. The learned attentions could capture distinct meaningful parts of the features for guiding the message passing. These attention maps are learned on the KITTI dataset for the task of monocular depth estimation.}
\label{fig:attention_kitti}
\vspace{-10pt}
\end{figure}

\par\noindent\textbf{End-to-End Network Optimization.} The parameters of the model consist of the front-end CNN parameters, 
$\vect{W}_c$, the parameters to produce the richer decomposition from each layer $l$, $\vect{W}_l$, the parameters of 
the AG-CRFs of the first level of the hierarchy, $\{\vect{W}_l^\textrm{I}\}_{l=1}^L$, and the parameters of the AG-CRFs of 
the second level of the hierarchy, $\vect{W}^\textrm{II}$. $L$ is the number of intermediate layers used from the 
front-end CNN. In order to jointly optimize all these parameters we adopt deep supervision~\cite{xie2015holistically} 
and we add an optimization loss associated to each AG-CRF module. 
We apply the proposed model into different pixel-wise prediction tasks, including contour detection, monocular depth estimation and semantic segmentation. For the contour detection task, as the contour detection problem is highly unbalanced, \ie~contour pixels are significantly less than non-contour pixels, we employ the modified 
cross-entropy loss function of~\cite{xie2015holistically}. Given a training data set ${\cal 
D}=\{(\vect{I}_p,\vect{E}_p)\}_{p=1}^P$ consisting of $P$ RGB-contour ground-truth pairs, the loss function $\ell$ 
writes:
\begin{equation}
\begin{aligned}
 \ell \big(\vect{W} \big) = & \sum_p \beta\sum_{e_p^k\in \vect{E}_p^{+}} \log \mathrm{P}\big(e^k_p=1 | 
\vect{I}_p; \vect{W} \big) +  \\ &\big(1-\beta\big) \sum_{e_p^k\in \vect{E}_p^{-}} \log \mathrm{P}\big(e^k_p=0 | 
\vect{I}_p; \vect{W} \big),
 \label{loss}
 \end{aligned}
 \end{equation} 
where $\beta =  |\vect{E}_p^{+}| / (|\vect{E}_p^{+}| + |\vect{E}_p^{-}|)$, $\vect{E}_p^{+}$ is the set of contour 
pixels of image $p$ and $\vect{W}$ is the set of all parameters. For the monocular depth estimation, we utilize an L2 loss for the continuous regression as in previous works~\cite{liu2015deep}, and for the semantic segmentation, we use a standard cross-entropy loss for multi-class classification as in~\cite{zhang2018context}. The network optimization is performed via the back-propagation algorithm with stochastic gradient descent.

\par\noindent\textbf{PGA-Net for pixel-wise prediction.} After training of the whole PGA-Net, the optimized network parameters 
$\vect{W}$ are used for the pixel-wise prediction task. Given a new test image $\vect{I}$, the $L+1$ classifiers produce a 
set of multi-scale prediction maps $\{\hat{\vect{E}}_l\}_{l=1}^{L+1} = \mathrm{PGA}\text{-}\mathrm{Net}(\vect{I}; 
\vect{W})$. Multi-scale predictions $\hat{\vect{E}}_l$ obtained from the AG-CRFs with elementary operations on the contour prediction task are shown in Fig.~\ref{predvis:bsds}.
We inspire from~\cite{xie2015holistically} to fuse the multiple scale predictions thus obtaining an average 
prediction $\hat{\vect{E}} = \sum_l \hat{\vect{E}}_l/(L+1)$. 

%
\section{Experiments}\label{sec:experiments}
We demonstrate the effectiveness of the proposed approach through extensive experiments on several publicly available benchmarks, and on three different tasks involving both the continuous domain (\ie~monocular depth estimation) and the discrete domain (\ie~object contour detection and semantic segmentation). We first introduce the experimental setup and then present our results and analysis. 


\subsection{Experimental Setup}
\subsubsection{Datasets.}
\par\noindent\textbf{BSDS500 and NYUD-V2 for object contour detection.} For the object contour detection task, we employ two different benchmarks: the {BSDS500} and the {NYUD-V2} datasets. The {BSDS500} dataset is an extended dataset based on BSDS300~\cite{arbelaez2011contour}. It consists of 200 training, 100 validation and 200 testing images. 
The groundtruth pixel-level labels for each sample are derived considering multiple annotators. Following~\cite{xie2015holistically,yang2016object}, we use all the training and validation images for learning the proposed model and perform data augmentation as described in~\cite{xie2015holistically}. 
The {NYUD-V2}~\cite{silberman2012indoor} contains 1449 RGB-D images and it is split into three subsets, consisting of 381 training, 414 validation and 654 testing images. 
Following~\cite{xie2015holistically} in our experiments we employ images at full resolution (\ie~$560 \times 425$ pixels) both in the training and in the testing phases. 

\par\noindent\textbf{KITTI for monocular depth estimation.} For the monocular depth estimation task, the {KITTI} dataset~\cite{Geiger2013IJRR} is considered. This dataset is collected for various important computer vision tasks within a context of self-driving. It contains depth video data capture using a LiDAR sensor installed on a driving car. To have a fair comparison with existing works, we follow a standard setting of the training and testing split originally proposed by Eigen~\etal~\cite{eigen2014depth}. There are in total 61 scenes selected from the raw data distribution. Specifically, we use total 22,600 frames from 32 scenes for training, and 697 frames from the rest 29 scenes for testing. The sparse ground-truth depth maps are obtained by reprojecting the 3D points captured from a velodyne laser onto the left monocular camera as in~\cite{garg2016unsupervised}. The resolution of RGB images are down-sampled to $621\times 188$ from the original resolution of $1224\times 368$ for training.

\par\noindent\textbf{Pascal-Context for semantic segmentation.} For the semantic segmentation task, we use the {Pascal-Context} dataset~\cite{mottaghi2014role}. The {Pascal-Context} dataset performs the augmentation of the pixel-level segmentation annotations on the Pascal VOC 2010, and enlarges the number of semantic classes from original 20 categories to more than 400 categories. Following previous works~\cite{chen2016deeplab, zhang2018context}, we evaluate on the setting with the most frequent 59 classes, in total 60 classes plus the background class. The rest classes are masked out during training.

\begin{table}[t]
\centering
\caption{BSDS500 dataset: quantitative results. \amid{All the compared methods use the official training and testing data. RCF~\cite{liu2016richer} utilizes extra training data for model learning. Our model outperforms Deep Crip Boundary~\cite{wang2018deep} considering the same ResNet50 backbone. Res16x is an enhanced backbone based on ResNet50.} \amid{3S/5S indicates three/five feature scales are considered for the structured fusion.}}
\vspace{-5pt}
\label{ods_bsd}
\resizebox{0.92\linewidth}{!} {
\begin{tabular}{l|c|ccc}
\toprule[0.3ex]
Method & \amid{Backbone} & ODS & OIS & AP \\\midrule
Human & - & .800 & .800     & - \\\midrule
Felz-Hutt\cite{felzenszwalb2004efficient} & -& .610 & .640 & .560\\
Mean Shift\cite{comaniciu2002mean} & - & .640 & .680 & .560 \\
Normalized Cuts\cite{shi2000normalized} & - & .641 & .674 & .447 \\
ISCRA\cite{ren2013image} & - & .724 & .752 & .783 \\
gPb-ucm\cite{arbelaez2011contour} & - & .726 & .760 & .727 \\
Sketch Tokens\cite{lim2013sketch} & - & .727 & .746 & .780 \\
MCG\cite{pont2015multiscale} & - & .747 & .779 & .759 \\
LEP\cite{zhaosegmenting} & - & .757 & .793 & .828 \\\midrule
DeepEdge~\cite{bertasius2015deepedge} & \amid{AlexNet} & .753 & .772 & .807 \\
DeepContour~\cite{shen2015deepcontour} & \amid{AlexNet} & .756 & .773 & .797 \\
HED~\cite{xie2015holistically} & \amid{VGG16} & .788 & .808 & .840 \\
CEDN~\cite{yang2016object} & \amid{VGG16} & .788 & .804 & .834 \\
COB~\cite{maninis2016convolutional} & \amid{ResNet50} & .793 & .820 & .859\\
\amid{{Deep Crip Boundary~\cite{wang2018deep}}} & \amid{ResNet50} & \textcolor{black}{.803} & \textcolor{black}{ .820} & \textcolor{black}{.871}\\
\amid{{Deep Crip Boundary~\cite{wang2018deep}}} & \amid{Res16x} & \textcolor{black}{.810} & \textcolor{black}{.829} & \textcolor{black}{.879}\\
\textcolor{gray}{RCF~\cite{liu2016richer} (not comp.)} & \amid{ResNet50} & \textcolor{gray}{.811} & \textcolor{gray}{.830} & \textcolor{gray}{--}\\
\midrule
PGA-Net (fusion) (3S) & \amid{ResNet50} & {.798} & {.829} & {.869} \\ 
PGA-Net (fusion) w/ CK (3S) & \amid{ResNet50} & {.799} & {.831} & {.872} \\ 
\amid{PGA-Net (fusion) w/ CK (5S)}& \amid{ResNet50} & \amid{\textbf{.805}} & \amid{\textbf{.835}} & \amid{\textbf{.878}} \\ 
\bottomrule[0.3ex]                       
\end{tabular}
}
\end{table}

\begin{figure}[!t]
\centering
\includegraphics[width=1\linewidth]{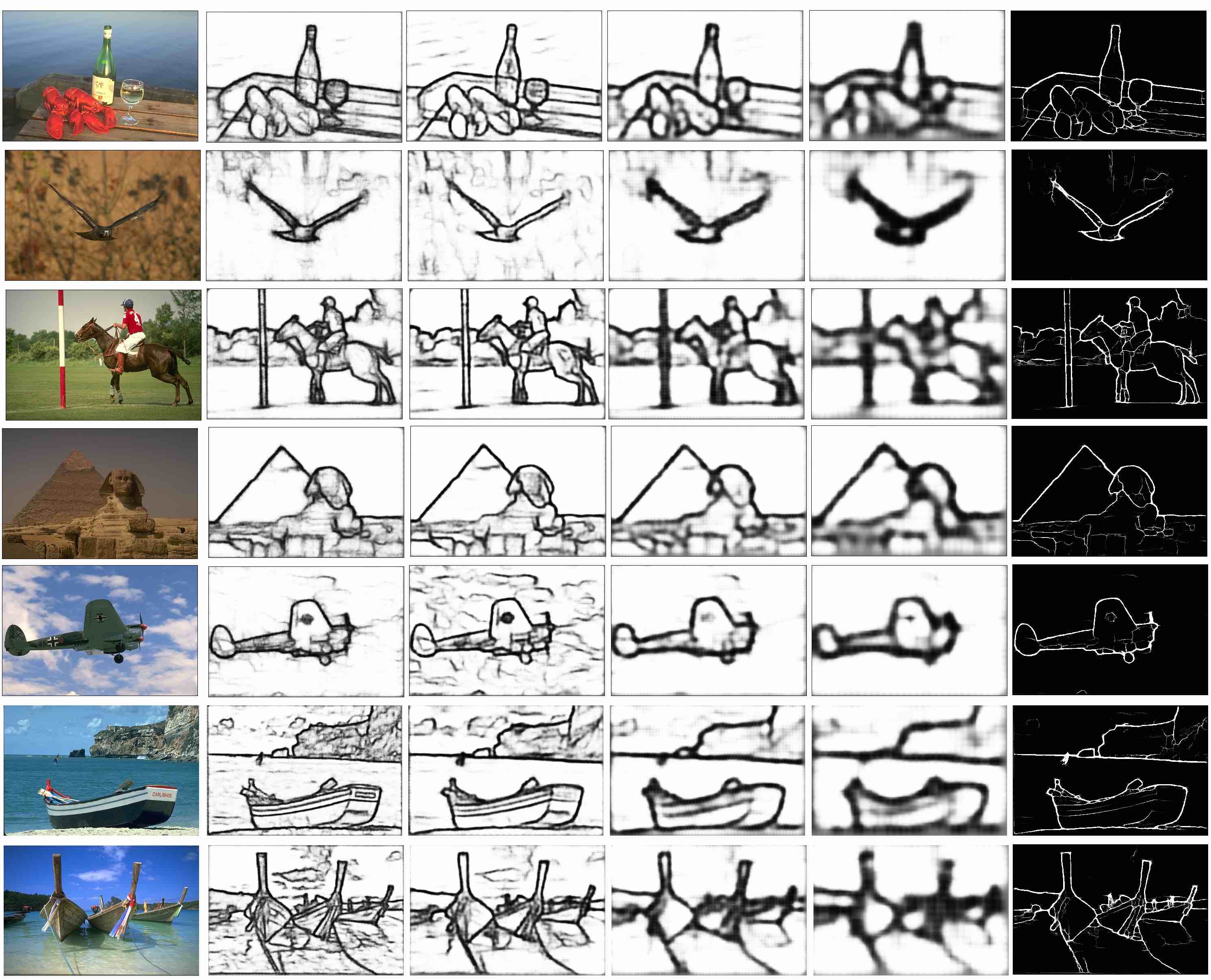}
\caption{Examples of predictions from different multi-scale features on BSDS500. The first column is the input test images. The 2nd to the 5nd columns show the predictions from different multi-scale features. The last column shows the final contour map after standard NMS.}
\label{predvis:bsds}
\end{figure}

\begin{figure*}[!t]
\centering
\subfigure[BSDS500]{\includegraphics[width=.445\linewidth]{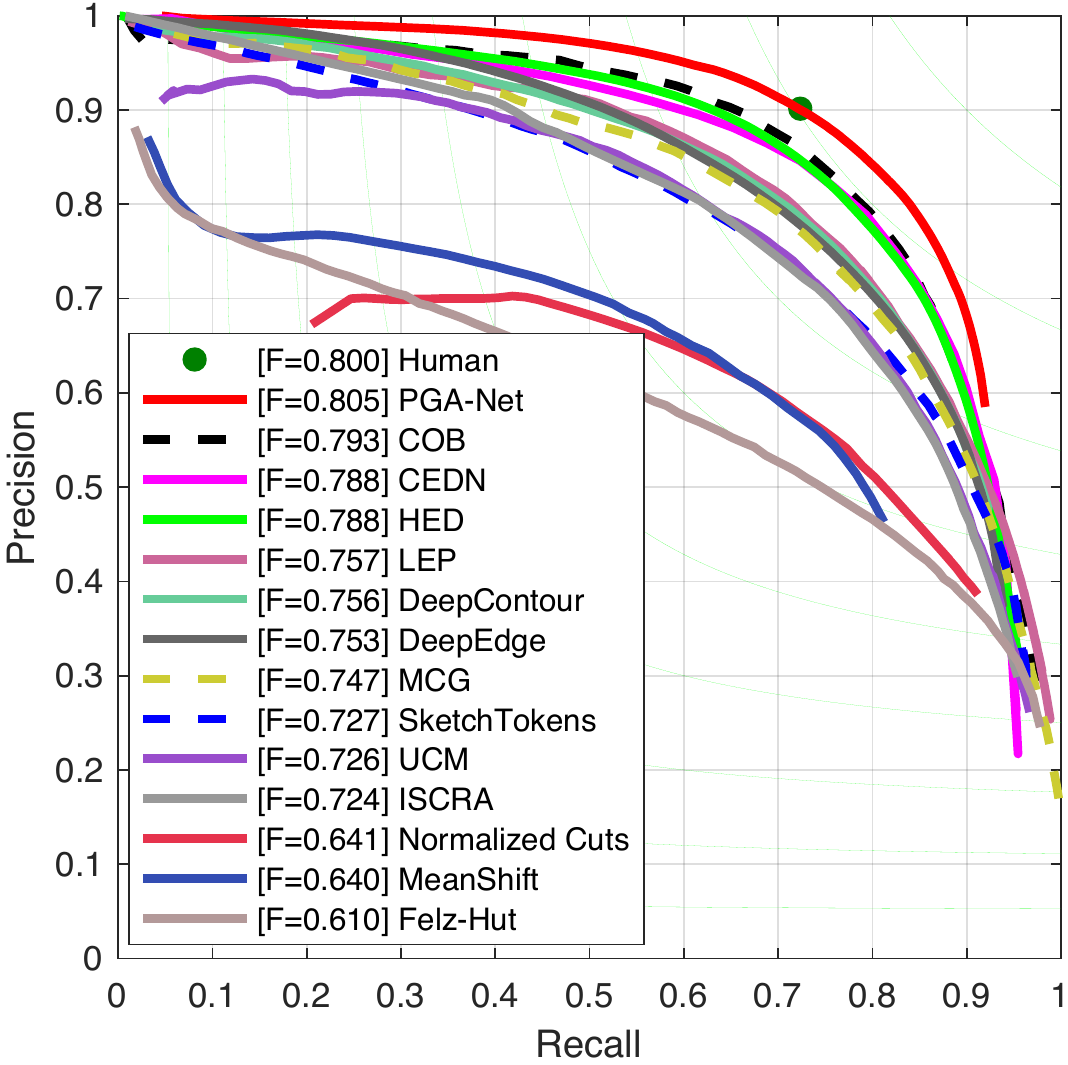}}
\hspace{0.6cm}
\subfigure[NYUD-V2]{\includegraphics[width=.445\linewidth]{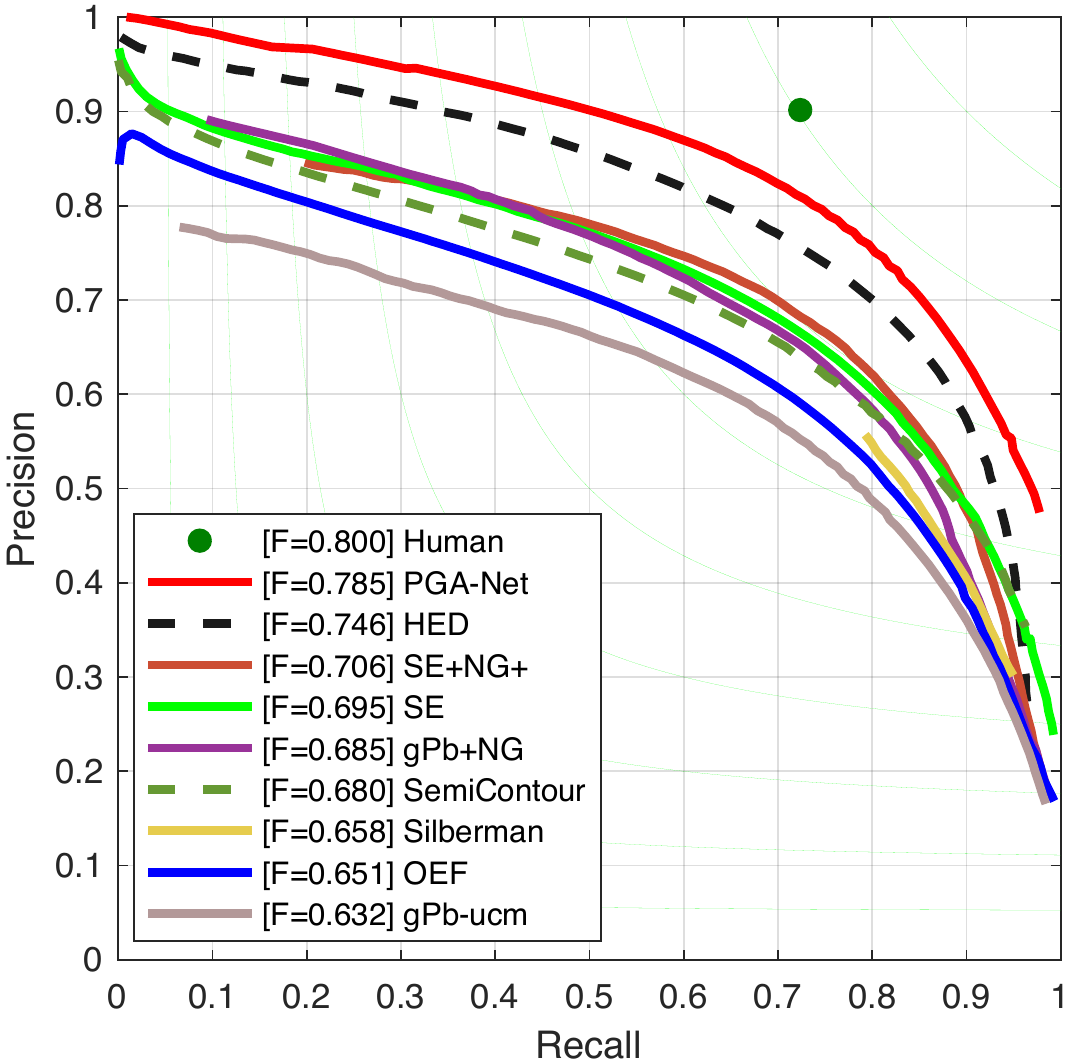}}
\vspace{-8pt}
\caption{Precision-Recall Curves on the BSDS500~\cite{arbelaez2011contour} and NYUD-V2~\cite{silberman2012indoor} test sets. The proposed PGA-Net achieves the best performance among the competitors on the ODS metric on both datasets. The results on NYUD-V2 are all based on the RGB and HHA data.}\label{fig:prcurves}
\vspace{-8pt}
\label{roc}
\end{figure*}

\subsubsection{Evaluation Metrics.}
\par\noindent\textbf{Evaluation protocol on object contour detection.} During the test phase standard non-maximum suppression (NMS)~\cite{dollar2013structured} is first applied to produce thinned contour maps. We then evaluate 
the detection performance of our approach according to different metrics, including the F-measure at Optimal Dataset Scale (ODS) and Optimal Image Scale (OIS) and the Average Precision (AP). The maximum tolerance allowed for correct matches of edge predictions to the ground truth is set to 0.0075 for the BSDS500 dataset, 
and to .011 for the NYUDv2 dataset as in previous works~\cite{dollar2013structured,gupta2014learning,xie2015holistically}.

\par\noindent\textbf{Evaluation protocol on monocular depth estimation.} Following the standard evaluation protocol as in previous works~\cite{eigen2015predicting,eigen2014depth,wang2015towards}, the following quantitative evaluation metrics are adopted in our experiments: 
\begin{itemize}[leftmargin=*]
\item mean relative error (rel): 
\( \frac{1}{K}\sum_{i=1}^K\frac{|\tilde{d}_i - d_i^\star|}{d_i^\star} \); 
\item root mean squared error (rms): 
\( \sqrt{\frac{1}{K}\sum_{i=1}^K(\tilde{d}_i - d_i^\star)^2} \);
\item mean log10 error (log10): \\
\( \frac{1}{K}\sum_{i=1}^K \Vert \log_{10}(\tilde{d}_i) - \log_{10}(d_i^\star) \Vert \);
\item scale invariant rms log error as used in~\cite{eigen2014depth}, rms(sc-inv.);
\item accuracy with threshold $t$: percentage (\%) of $d_i^\star$, \\subject to $\max (\frac{d_i^\star}{\tilde{d}_i}, \frac{\tilde{d}_i}{d_i^\star}) = 
\delta < t~(t \in [1.25, 1.25^2, 1.25^3])$.
\end{itemize}
Where $\tilde{d}_i$ and $d_i^\star$ is the ground-truth depth and the estimated depth at pixel $i$ respectively; $K$ is the total number of pixels of the test images. 

\par\noindent\textbf{Evaluation protocol on semantic segmentation.} Following previous works and use the DeepLab evaluation tool, we report our quantitative results on the standard metrics of pixel accuracy (pixAcc) and mean intersection over union (mIoU) averaged over classes. Both metrics are the higher the better. The background category is all included in the evaluation as in previous works~\cite{zhou2017scene, zhang2018context}.

\subsubsection{Implementation Details.} 
The proposed PGA-Net is implemented under the deep learning framework~\textit{Pytorch}. 
The training and testing phase are carried out on four Nvidia Tesla P40 GPUs, each with 24GB memory. The ResNet-50 and ResNet-101 networks pretrained on ImageNet~\cite{deng2009imagenet} are used to initialize the front-end CNN of PGA-Net for different backbone experiments. To consider the computational efficiency, our implementation only employs three scales, \ie~we generate multi-scale features from three different semantic layers of the 
backbone CNN (\ie~\emph{res3d}, \emph{res4f}, \emph{res5c} in ResNet). In our CRF model we consider dependencies between all scales. 
Within the AG-CRFs, the kernel size for all convolutional operations is 
set to $3\times 3$ with stride $1$ and padding $1$. 
The weighting parameters $a^i_{s_r}$ are learned automatically via using convolutional operations with a kernel size of $1\times 1$.
For the object contour detection, the initial learning rate is set to 1e-7 in all our experiments, and decreases 10 times after every 10k iterations. The total number of iterations for BSDS500 and NYUD v2 is 40k and 30k, respectively. The momentum and weight decay parameters are set to 0.9 and 0.0002, as in~\cite{xie2015holistically}. As the training images
have different resolution, we need to set the batch size to 1, and for the sake of smooth convergence we updated the parameters only every 10 
iterations. \amid{For the monocular depth estimation and the semantic segmentation task, following previous works~\cite{godard2016unsupervised, xu2018monocular, zhang2018context} for fair comparison, the batch size is set to 8 and 16, respectively;} the learning rate is set to 0.001 with a momentum of 0.9 and a weight decay of 0.0001 using a polynomial learning rate scheme as used in~\cite{zhang2018context,chen2016deeplab}. \amid{Regarding the overall training time, it takes around 8, 13, 20 hours for the contour detection on BSDS500, depth estimation on KITTI and semantic segmenation on Pascal-Context. Our model also achieves almost real-time inference time (around 8 frames per second) for all the three tasks.}


\begin{table}[t]
\centering
\caption{Quantitative performance comparison on NYUD-V2 RGB dataset for the contour detection task with the official training/testing protocols. \amid{We achieve better performance in terms of all the metrics than RCF~\cite{liu2016richer}, which is the best performing method on BSDS500 with extra training data.} \amid{3S/5S indicates three/five feature scales are considered.}}
\vspace{-5pt}
\label{ods_nyu}
\resizebox{0.92\linewidth}{!} {
\begin{tabular}{l|c|ccc}
\toprule[0.3ex]
Method & \amid{Backbone} & ODS & OIS & AP \\\midrule
gPb-ucm~\cite{arbelaez2011contour} & -- & .632 & .661 & .562 \\
OEF~\cite{hallman2015oriented} & -- & .651 & .667 & -- \\
Silberman \etal~\cite{silberman2012indoor} & -- & .658 & .661 & -- \\
SemiContour~\cite{zhang2016semicontour} & -- & .680 & .700 & .690 \\
SE~\cite{dollar2015fast} & -- & .685 & .699 & .679 \\
gPb+NG~\cite{gupta2013perceptual} & -- & .687 & .716 & .629 \\
SE+NG+~\cite{gupta2014learning} & -- & .710 & .723 & .738 \\ \midrule
HED (RGB)~\cite{xie2015holistically} & \amid{VGG16} & .720 & .734 & .734 \\
HED (HHA)~\cite{xie2015holistically} & \amid{VGG16} & .682 & .695 & .702 \\
HED (RGB + HHA)~\cite{xie2015holistically} & \amid{VGG16} & .746 & .761 & .786  \\ 
RCF (RGB) + HHA)~\cite{liu2016richer} & \amid{VGG16} & .757 &.771 & -- \\
RCF (RGB) + HHA)~\cite{liu2016richer} & \amid{ResNet50} & {.781} &.793 & -- \\\midrule
PGA-Net (HHA) (3S) & \amid{ResNet50} & .716 & .729 & .734 \\
PGA-Net (RGB) (3S) & \amid{ResNet50} & .744 & .758 & .765 \\
PGA-Net (RGB+HHA) (3S) & \amid{ResNet50} & {.771} & {.786} & {.802} \\
PGA-Net (RGB+HHA) w/ CK (3S) & \amid{ResNet50} & {.780} & {.795} & {.813} \\
\amid{PGA-Net (RGB+HHA) w/ CK (5S)} & \amid{ResNet50} & \amid{\textbf{.785}} & \amid{\textbf{.799}} & \amid{\textbf{.816}} \\
\bottomrule[0.3ex]            
\end{tabular}
}
\vspace{-15pt}
\end{table}

\subsection{Experimental Results}
In this section, we present the results of our evaluation, comparing the proposed model with several state of the art methods respectively on the three different tasks. We further conduct an in-depth analysis of our model,
to show the impact of different components on the performance. And Finally we present some qualitative results and analysis of the model.
\subsubsection{Comparison with state of the art methods.} 
\par\textbf{Comparison on BSDS500 and NYUD-V2.} We first consider the BSDS500 dataset and compare the performance of our approach with several traditional contour detection methods, including Felz-Hut~\cite{felzenszwalb2004efficient}, MeanShift~\cite{comaniciu2002mean}, 
Normalized Cuts~\cite{shi2000normalized}, ISCRA~\cite{ren2013image}, 
gPb-ucm~\cite{arbelaez2011contour}, SketchTokens~\cite{lim2013sketch}, MCG~\cite{pont2015multiscale}, LEP~\cite{zhaosegmenting}, and more recent CNN-based 
methods, including DeepEdge~\cite{bertasius2015deepedge}, DeepContour~\cite{shen2015deepcontour}, HED~\cite{xie2015holistically}, CEDN~\cite{yang2016object}, 
COB~\cite{maninis2016convolutional}, \amid{and Deep Crisp Boundaries~\cite{wang2018deep}}. We also report results of the RCF method~\cite{liu2016richer}, although they are not comparable because in~\cite{liu2016richer} an extra dataset (\ie~Pascal Context) \amid{which is even larger than BSDS500, was used during RCF training to improve the results on BSDS500.}
In this series of experiments we consider PGA-Net with FLAG-CRFs.
The results of this comparison are shown in Table~\ref{ods_bsd} and Figure~\ref{roc}a. 
PGA-Net obtains an F-measure (ODS) of 0.798, thus outperforms all previous methods. The improvement over the second and third best approaches, \ie~COB and HED,
is 0.5\% and 1.0\%, respectively, which is not trivial to achieve on this challenging dataset. Furthermore, when considering the OIS and AP metrics, our approach is also better, with a clear performance gap. By using the proposed strategy of conditional kernels, we further clearly boost the performance of PGA-Net on all the three metrics (\ie~ODS, OIS and AP), on this performance saturated dataset. \amid{Besides, comparing to Deep Crisp Boundaries, ours using 3 feature scales is comparable on the ODS metric while achieves better performance w.r.t. both the OIS and AP metrics if the same backbone architecture (\ie~ResNet50) is considered for both methods. While five feature scales are used for the structured fusion, ours outperforms Deep Crisp Boundaries on all the metrics with the ResNet50 backbone.}
\par To conduct the experimental comparison on NYUDv2, following previous works~\cite{xie2015holistically} we also consider three different types of input representations,~\ie~RGB, HHA \cite{gupta2014learning}
and RGB-HHA data. 
The HHA data~\cite{gupta2014learning} are encoded depth feature images, in which the three image chanels are horizontal disparity, 
height above ground, and angle of the pixel's local surface normal with the inferred direction of gravity, respectively. 
The results corresponding to the use of both RGB and HHA data (\ie~RGB+HHA) are obtained by performing a weighted average of the estimates obtained from two PGA-Net models trained separately on RGB and HHA representations. As baselines we 
consider gPb-ucm~\cite{arbelaez2011contour}, OEF~\cite{hallman2015oriented}, the method in~\cite{silberman2012indoor}, 
SemiContour~\cite{zhang2016semicontour}, SE~\cite{dollar2015fast}, gPb+NG~\cite{gupta2013perceptual}, SE+NG+~\cite{gupta2014learning},  
HED~\cite{xie2015holistically} and RCF~\cite{liu2016richer}. \amid{On the NYUD-V2 dataset, our approach outperforms the RCF~\cite{liu2016richer} which obtained the best performance on BSDS500 with extra training data, with a clear performance gap, where the experimental protocol for both is exactly the same on this dataset.}
\begin{table}[!t]
\centering
\caption{Quantitative performance analysis of the proposed PGA-Net on NYUD-V2 RGB dataset for the contour detection task. \amid{H2 indicates only the second hierarchy is utilized; 3S and 5S indicate three and five feature scales are considered for the structured fusion, respectively.}}
\vspace{-5pt}
\label{nyud-ablation}
\resizebox{0.95\linewidth}{!} {
\begin{tabular}{l|c|ccc}
\toprule[0.3ex] 
Method & Backbone & ODS & OIS & AP \\\midrule
\amid{Hypercolumn~\cite{hariharan2015hypercolumns} (5S)} & \amid{ResNet50} & \amid{.720} & \amid{.731} & \amid{.733} \\
\amid{HED~\cite{xie2015holistically} (5S)} & \amid{ResNet50} & \amid{.722} & \amid{.737} & \amid{.738} \\\midrule
PGA-Net (baseline) (H2, 3S) & \amid{ResNet50} & .711 & .720 & .724 \\
\amid{PGA-Net (w/o AG-CRFs) (3S)} & \amid{ResNet50} & \amid{.722} & \amid{.732} & \amid{.739}\\
PGA-Net (w/ CRFs) (3S) & \amid{ResNet50} & .732 & .742 & .750 \\
PGA-Net (w/o deep supervision) (3S) & \amid{ResNet50} & .725 & .738 & .740  \\
\midrule
\amid{PGA-Net (w/ PLAG-CRFs) (H2, 3S)} & \amid{ResNet50} & \amid{.731} & \amid{.742} & \amid{.743} \\
PGA-Net (w/ PLAG-CRFs) (3S) & \amid{ResNet50} & .737 & .749 & .746 \\
PGA-Net (w/ FLAG-CRFs) (3S) & \amid{ResNet50} & {.744} & {.758} & {.765} \\ 
PGA-Net (w/ FLAG-CRFs + CK) (3S) & \amid{ResNet50} & {.751} & {.767} & {.778} \\ 
\amid{PGA-Net (w/ FLAG-CRFs + CK) (5S)} & \amid{ResNet50} & \amid{\textbf{.754}} & \amid{\textbf{.772}} & \amid{\textbf{.781}} \\
\bottomrule[0.3ex]                        
\end{tabular}
}
\vspace{-5pt}
\end{table}
All of them are reported in Table~\ref{ods_nyu} and Figure~\ref{roc}b. Again, our final model (PGA-Net w/ CK) significantly outperforms all previous comparison methods. In particular, the increased performance with respect to HED~\cite{xie2015holistically} and RCF~\cite{liu2016richer} confirms the benefit of the proposed multi-scale feature learning
and fusion scheme. 

\begin{figure*}[!t]
\centering
\includegraphics[width=0.99\linewidth]{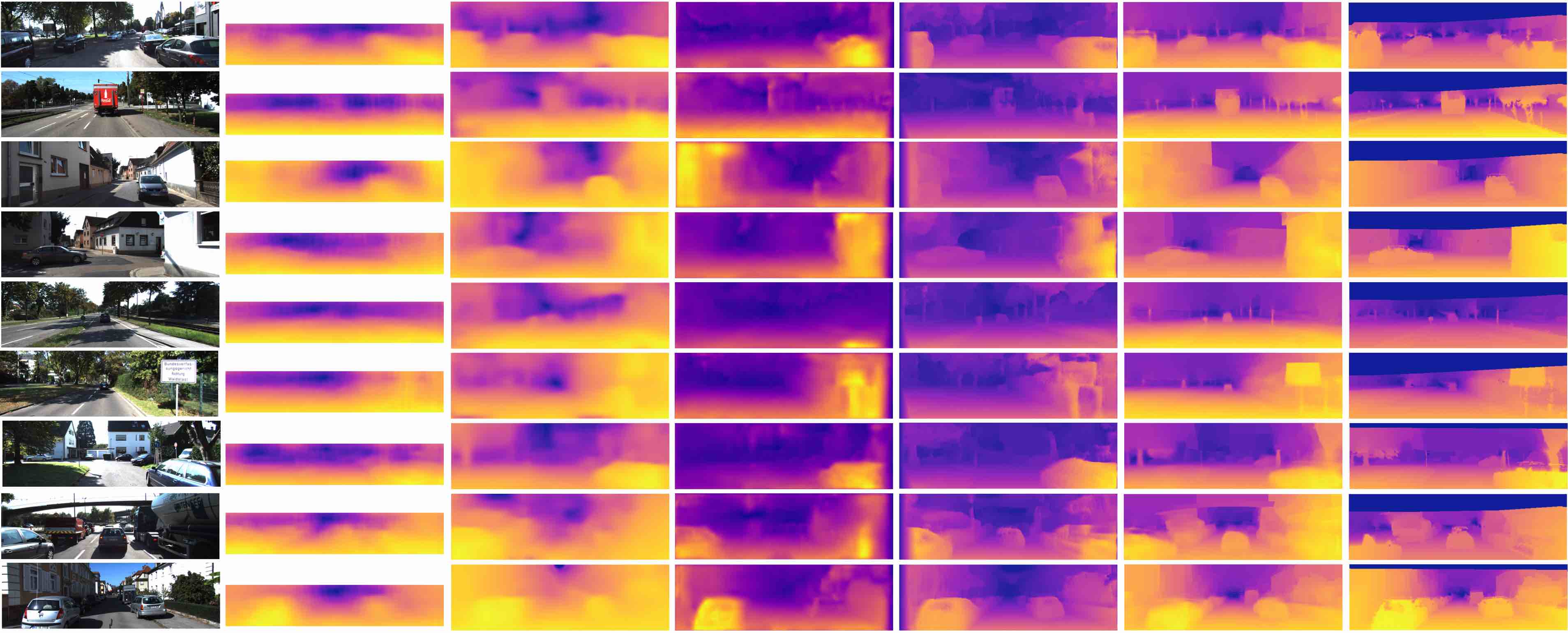} 
\put(-493,210){\footnotesize RGB Image}
\put(-428,210){\footnotesize Eigen~\etal~\cite{eigen2014depth}}
\put(-352,210){\footnotesize Zhou~\etal~\cite{zhou2017unsupervised}}
\put(-280,210){\footnotesize Garg~\etal~\cite{garg2016unsupervised}}
\put(-212,210){\footnotesize Godard~\etal~\cite{godard2016unsupervised}}
\put(-123,210){\footnotesize Ours }
\put(-63,210){\footnotesize GT Depth Map}
\vspace{-8pt}
\caption{Qualitative examples of monocular depth prediction on the KITTI raw dataset. The comparison with other competive methods including Eigen~\etal~\cite{eigen2014depth}, Zhou~\etal~\cite{zhou2017unsupervised}, Garg~\etal~\cite{garg2016unsupervised} and Godard~\etal~\cite{godard2016unsupervised} are presented. We perform bilinear interpolation on the sparse ground-truth depth maps for better visualization.}
\label{predvis:kitt}
\vspace{-10pt}
\end{figure*}
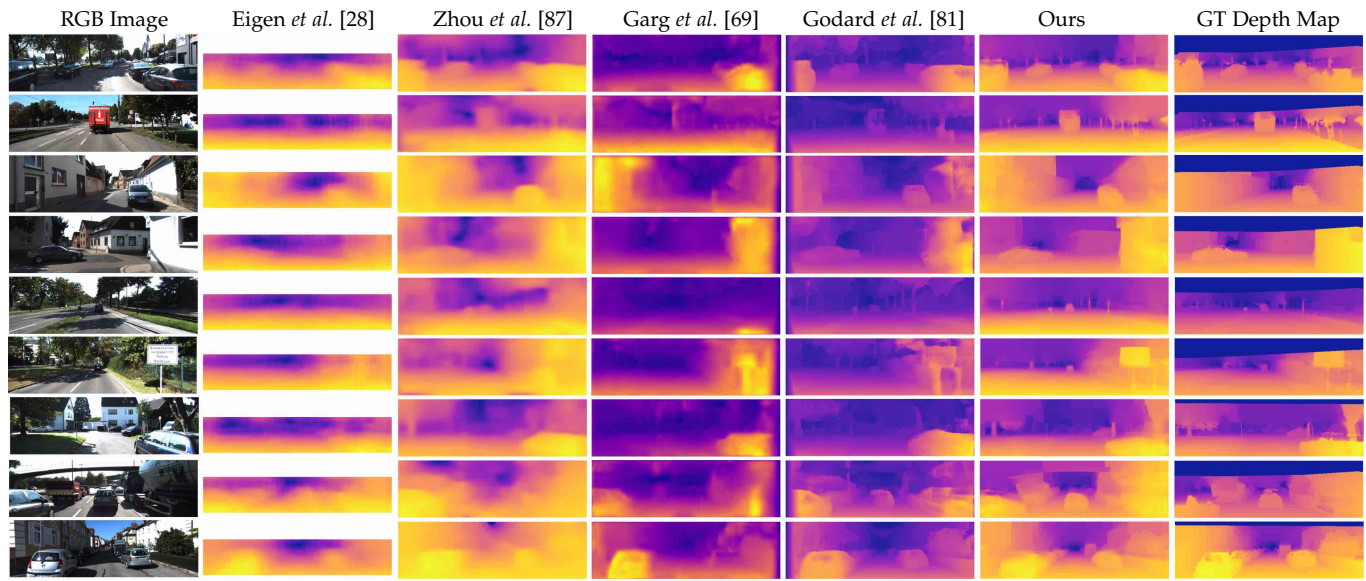

\begin{table*}[!t]
\centering
\caption{Quantitative comparison with the state of the art methods on the KITTI raw dataset for monocular depth estimation. The proposed PGA-Net achieves top performance over all the competitors w.r.t. all the evaluation metrics. The standard training and testing sets split by Eigen~\etal~\cite{eigen2014depth} are used. The `range' means different ground-truth depth range for evaluation, and the 'sup' means the ground-truth depth is used for supervision in the training. `CK' denotes the proposed conditional kernel strategy. The methods requiring video data are marked with $^*$. The DORN method needs to assume the depth range in the training, which is not the same as our continuous regression setting, and thus not directly comparable, and we are highly complementary to each other since we work on learning effective representation while theirs focus on the loss level. \amid{3S/5S indicates three/five feature scales from the front-end CNN are considered for the structured fusion by the proposed model.}}
\vspace{-5pt}
\setlength\tabcolsep{8pt}
\resizebox{0.97\linewidth}{!} {
\begin{tabular}{l|cc|cccc|ccc}
\toprule[0.3ex]
\multirow{2}{*}{Method} & \multicolumn{2}{c|}{Setting} & \multicolumn{4}{c|}{\tabincell{c}{Error (lower is better)}} & \multicolumn{3}{c}{\tabincell{c}{Accuracy (higher is better)}} \\\cline{2-10}
                                      & range & sup? & rel & sq rel & rmse & rmse (log) & $\delta < 1.25$ & $\delta < 1.25^2$ & $\delta < 1.25^3$ \\\midrule
Garg~\etal~\cite{garg2016unsupervised}  & 80m & No & 0.177 & 1.169  & 5.285   & -  &  0.727 & 0.896  &  0.962   \\
Garg~\etal~\cite{garg2016unsupervised} L12 + Aug 8x  & 50m & No & 0.169  & 1.080   & 5.104  & -  &  0.740 & 0.904  &  0.958   \\
Godard~\etal~\cite{godard2016unsupervised}  & 80m & No & 0.148 & 1.344  & 5.927 & 0.247  &  0.803 & 0.922  &  0.964   \\
Zhou$^*$~\etal~\cite{zhou2017unsupervised} & 80m & No
& 0.208   &  1.768  & 6.858  &  0.283 & 0.678   & 0.885   & 0.957  \\
AdaDepth~\cite{adadepth} & 50m & No & 0.203 & 1.734 & 6.251 & 0.284 & 0.687 & 0.899 & 0.958 \\ 
Pilzer~\etal~\cite{pilzer2018unsupervised} & 80m & No & 0.152 &
 1.388 & 6.016 & 0.247  & 0.789 & 0.918 & 0.965 \\
Wang \& Lucey~\etal~\cite{wang2017learning} & 80m & No & 0.151 &
 1.257 & 5.583 & 0.228  & 0.810 & 0.936 & 0.974 \\ 
DF-Net$^*$~\cite{zou2018df} & 80m & No & 0.150 & 1.124 & 5.507 & 0.223 & 0.806 & 0.933 & 0.973 \\
Zhan$^*$~\etal~\cite{zhan2018unsupervised} & 80m & No & 0.144 & 1.391 & 5.869 & 0.241  & 0.803 & 0.933 & 0.971 \\
Kuznietsov \etal \cite{kuznietsov2017semi}  & 80m & No & - & - & {4.621} & -  & {0.852} & {0.960} & {0.986}  \\
\midrule
Saxena~\etal~\cite{saxena2009make3d}   & 80m & Yes & 0.280  &  - & 8.734 & 0.327 & 0.601  &  0.820  &  0.926\\
Liu~\etal~\cite{liu2015deep}    & 80m & Yes & 0.217  &  0.092   &  7.046 & - &  0.656 &  0.881  &  0.958 \\
Eigen~\etal~\cite{eigen2014depth}    & 80m & Yes & 0.190  &  -      &  7.156 & 0.246  &  0.692 &  0.899  &  0.967 \\
Mahjourian$^*$~\cite{mahjourian2018unsupervised} & 80m & Yes & 0.163 & 1.240 & 6.220 &  0.250 & 0.762 & 0.916 & 0.968 \\
MS-CRF~\cite{xu2018monocular} & 80m & Yes & 0.125 & 0.899 & 4.685 & - & 0.816 & 0.951 & 0.983 \\
\amid{Kuznietsov~\etal~\cite{kuznietsov2017semi} (supervised \& stereo)} & \amid{80m} & \amid{Yes} & \amid{-} & \amid{-} & \amid{4.621} & \amid{0.189} & \amid{0.862} & \amid{0.960} & \amid{0.986} \\
\amid{Gan~\etal~\cite{gan2018monocular}} & \amid{80m} & \amid{Yes} & \amid{0.098} & \amid{0.666} & \amid{3.933} & \amid{0.173} & \amid{0.890} & \amid{0.964} & \amid{0.985} \\
\amid{DORN~\cite{fu2018deep}} & \amid{{80m}} & \amid{{Yes}} & \amid{{0.072}} & \amid{{0.307}} & \amid{{2.727}} & \amid{{0.120}} & \amid{{0.932}} & \amid{{0.984}} & \amid{{0.994}} \\
\amid{Lee~\etal~(ResNet)~\cite{lee2019big}} & \amid{80m} & \amid{Yes} & \amid{0.061} & \amid{0.261} & \amid{2.834} & \amid{0.099} & \amid{0.954} & \amid{0.992} & \amid{0.998} \\
\midrule\midrule 
PGA-Net (baseline) &80m & Yes & 0.152  & 0.973  & 4.902  & 0.176 & 0.782 & 0.931 & 0.975 \\
PGA-Net (w/ CRFs) (3S)  & 80m & Yes & 0.140 & 0.942 & 4.823 & 0.171  & 0.793  & 0.941 &  0.979\\
PGA-Net (w/ PLAG-CRFs) (3S)  & 80m & Yes & 0.134 & 0.909 & 4.796 & 0.167  & 0.801  & 0.951 &  0.981\\
PGA-Net (w/ FLAG-CRFs) (3S) & 80m & Yes & {0.126}  & {0.901} & {4.689} & {0.157} & 0.813  & {0.950} & {0.982} \\
PGA-Net (w/ FLAG-CRFs + CK) (3S)& 80m & Yes & {0.118}  & {0.752} & {4.449}  & {0.181} & {0.852} & {0.962} & {0.987}\\
\amid{PGA-Net (w/ FLAG-CRFs + CK + DORN) (3S)}& \amid{80m} & \amid{Yes} & \amid{{0.063}}  & \amid{{0.267}} & \amid{{2.634}}  & \amid{{0.101}} & \amid{{0.952}} & \amid{{0.992}} & \amid{{0.998}}\\
\amid{PGA-Net (w/ FLAG-CRFs + CK + DORN) (5S)}& \amid{80m} & \amid{Yes} & \amid{\textbf{0.060}}  & \amid{\textbf{0.258}} & \amid{\textbf{2.595}}  & \amid{\textbf{0.097}} & \amid{\textbf{0.954}} & \amid{\textbf{0.993}} & \amid{\textbf{0.998}}\\
\bottomrule[0.3ex]                           
\end{tabular}
}
\label{overall_KITTI}
\vspace{-10pt}
\end{table*}

\par\noindent\textbf{Comparison on KITTI.} 
The state of the art comparison on the KITTI dataset for monocular depth estimation is shown in Table~\ref{overall_KITTI}. We compare with the methods with both supervised and unsupervised settings. For the unsupervised setting, the representative works such as Zhou~\etal~\cite{zhou2017unsupervised}, Garg~\etal~\cite{garg2016unsupervised}, Godard~\etal~\cite{godard2016unsupervised},
and Kuznietsov~\etal~\cite{kuznietsov2017semi} are compared. For the supervised methods, we consider 
the very competitive works such as Eigen~\etal~\cite{eigen2014depth}, Liu~\etal~\cite{liu2015deep}, Kuznietsov~\etal~\cite{kuznietsov2017semi}, \amid{Gan~\etal~\cite{gan2018monocular}}, DORN~\cite{fu2018deep} \amid{and Lee~\etal~\cite{lee2019big}}. 
Our approach also employs the supervised setting using single monocular images in training and testing. 
As shown in Table~\ref{overall_KITTI}, \amid {our approach achieves top-level performance compared with both the supervised and unsupervised comparison methods.} Specifically, DORN needs to assume the depth range in the training, which is not the same setting as our continuous regression, and thus not directly comparable to ours. Besides, DORN specifically works on the predictions via using an ordinal regression loss, while ours focuses on learning effective representations, therefore we are complementary to each other. \amid{By combing the proposed AG-CRF module with DORN, specifically with its multi-scle backbone and ordinary regression module, we achieve clearly better performance than DORN, which further confirms the effectiveness of the proposed AG-CRF model. Our approach also obtains the same level of performance compared with the best performing method Lee~\etal~\cite{lee2019big}~using 3 feature scales. We outperform Lee~\etal~\cite{lee2019big}~while 5 feature scales are further considered.} More importantly, the proposed graph-based approach obtains significantly better results than the CRF-based methods (\ie~MS-CRF~\cite{xu2018monocular} and Liu~\etal~\cite{liu2015deep}) on the monocular depth estimation task.

\par\noindent\textbf{Comparison on Pascal-Context.} We compare the proposed PGA-Net with the most competitive methods on the Pascal-Context dataset, including ASPP~\cite{chen2017rethinking}, PSPNet~\cite{zhao2016pyramid}, EncNet~\cite{zhang2018context} and HRNet~\cite{sun2019high}. The experiments are conducted on both ResNet-50 and ResNet-101 backbone networks. Our PGA-Net is 2.24 and 3.05 points better on the mIoU metric than the popular method (\ie~EncNet, which considers a channel-wise attention for feature refinement) with ResNet-50 and ResNet-101 respectively. \amid{Compared with another multi-scale method HRNet, which utilizes the multi-scale aggregation from feature maps with different resolutions to boost the final performance, our PGA-Net obtains better performance with a clear gap. Note that the backbone structure HRNetV2-W48 used by HRNet has a bigger network capacity than the ResNet101 backbone we used. Our approach also obtains the same level performance comparing with the best performing method OCR considering the same backbone ResNet101. By using five feature scales for the structured fusion in the proposed model, our model outperforms OCR (55.1 vs. 54.8 in terms of mIoU using the same ResNet101 backbone). More importantly, the core idea of OCR utilizing the soft object regions is complementary to ours.}  We could also observe that our graph-based method significantly outperforms deep-lab-v2~\cite{chen2016deeplab} which also utilizes a CRF model.

\subsubsection{Model Analysis.} 
\par\noindent\textbf{Baseline models.} To further demonstrate the effectiveness of the proposed model and
analyze the impact of the different components of PGA-Net on the countour detection task, we conduct an ablation study considering the NYUDv2 (RGB data) and the Pascal-Context dataset. We evaluated the following baseline models: (i) PGA-Net (baseline), which removes the first-level hierarchy and directly 
concatenates the feature maps for prediction, (ii) PGA-Net (w/o AG-CRFs), which employs the proposed multi-scale hierarchical structure but 
discards the AG-CRFs, (iii) PGA-Net (w/ CRFs), which replaces our AG-CRFs with a multi-scale CRF model without attention gating, (iv) PGA-Net (w/o deep supervision) obtained by removing intermediate loss functions in PGA-Net, (v) PGA-Net with the proposed two versions of the AG-CRFs model, \ie~PLAG-CRFs and FLAG-CRFs, and (vi) PGA-Net w/ CK, which uses the proposes conditional kernel strategy. We also consider as reference traditional multi-scale
deep learning models employing multi-scale representations, \ie~Hypercolumn~\cite{hariharan2015hypercolumns} and
HED~\cite{xie2015holistically}.

\begin{figure}[!t]
\centering
\includegraphics[width=0.9\linewidth]{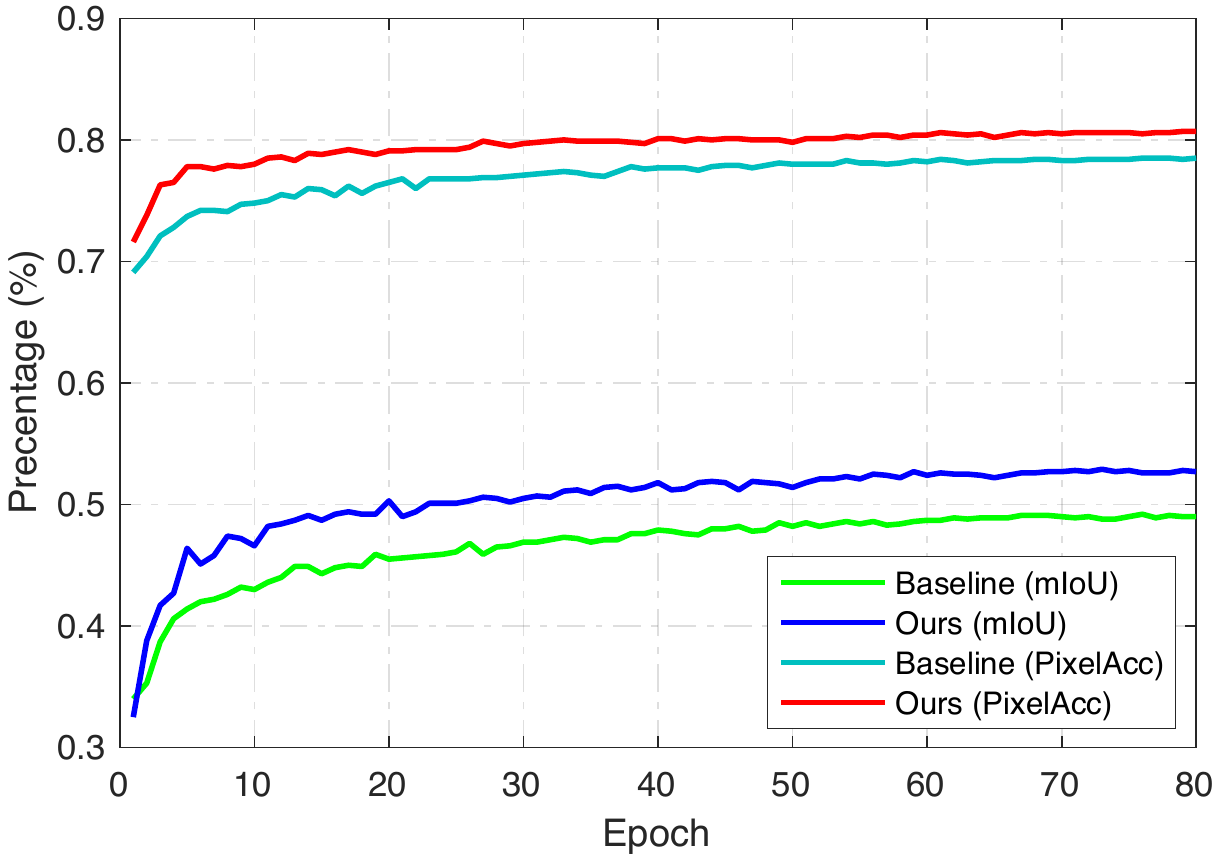}
\vspace{-10pt}
\caption{Training curves of our approach and the baseline model \amid{(\ie~ours with the proposed AG-CRFs disabled)} in terms of both the mIoU and PixAcc metrics on the Pascal-Context dataset. The number of overall training epochs is 80.}
\vspace{-5pt}
\label{fig:comp_miou_pixelacc}
\end{figure}

\begin{table}[!t]
\centering
\caption{Quantitative performance analysis of the proposed PGA-Net with the ResNet 50 backbone on Pascal-Context for the semantic segmentation task. `CK' denotes the proposed conditional kernel strategy. \amid{H2 denotes only the second hierarchy is used. 3S, 4S and 5S indicate three, four and five feature scales from different stages are considered, respectively.}}
\vspace{-5pt}
\label{pascal-ablation}
\resizebox{0.99\linewidth}{!} {
\begin{tabular}{l|c|cc}
\toprule[0.3ex] 
Method & \amid{Backbone} & pixAcc(\%) &  mIoU(\%) \\\midrule
Hypercolumn~\cite{hariharan2015hypercolumns} (5S) & \amid{ResNet50} & 75.96 & 47.88 \\
HED~\cite{xie2015holistically} (5S) & \amid{ResNet50} & 76.45 & 48.41
\\
\amid{HRNet~\cite{sun2019high}} & \amid{HRNetV2-W48} & \amid{-} & \amid{54.00} \\\midrule
PGA-Net (baseline) (H2, 3S) & \amid{ResNet50} & 75.60 & 47.21 \\
PGA-Net (w/o AG-CRFs) (3S)& \amid{ResNet50} & 76.70 & 48.73\\
PGA-Net (w/ CRFs) (3S) & \amid{ResNet50} & 77.10 & 49.12 \\
PGA-Net (w/o deep supervision) (3S) & \amid{ResNet50} & 76.90 & 48.92  \\\midrule
\amid{PGA-Net (w/ PLAG-CRFs) (H2, 3S)} & \amid{ResNet50} & \amid{77.41} & \amid{49.22} \\
PGA-Net (w/ PLAG-CRFs) (3S) & \amid{ResNet50} & 77.91 & 50.13 \\
PGA-Net (w/ FLAG-CRFs) (3S) & \amid{ResNet50} & {78.49} & {51.01} \\
PGA-Net (w/ FLAG-CRFs + CK) (3S) & \amid{ResNet50} & \textbf{79.62} & \textbf{52.15} \\ 
\amid{PGA-Net (w/ FLAG-CRFs + CK) (5S)} & \amid{ResNet50} & \amid{\textbf{80.42}} & \amid{\textbf{53.25}} \\
\amid{PGA-Net (w/ FLAG-CRFs + CK) (4S)} & \amid{HRNetV2-W48} & \amid{\textbf{81.31}} & \amid{\textbf{55.18}} \\
\bottomrule[0.3ex]                  
\end{tabular}
}
\vspace{-8pt}
\end{table}

\par\noindent\textbf{Analysis.} The quantitative results on different baseline models are shown in Table~\ref{nyud-ablation} and~\ref{pascal-ablation}. The results clearly show the advantages of our contributions. The ODS F-measure of PGA-Net (w/o AG-CRFs) is 1.1\% 
higher than PGA-Net (baseline), 
clearly demonstrating the effectiveness of the proposed hierarchical network and confirming our intuition that exploiting more 
richer and diverse multi-scale representations is beneficial, which could be also verified from the results on the Pascal-Context as shown in Table~\ref{pascal-ablation}.  
Table~\ref{nyud-ablation} also shows that our AG-CRFs plays a fundamental role for accurate detection, as PGA-Net (w/ FLAG-CRFs) leads to an improvement of 1.9\%
over PGA-Net (w/o AG-CRFs) in terms of OSD. \amid{Besides, we could also observe that the PGA-Net (w/ PLAG-CRF) also boosts the performance over the PGA-Net (baseline) (.711 vs. .731 in terms of the ODS metric) where both of them use only the second hierarchy.} Finally, PGA-Net (w/ FLAG-CRFs) is 1.2\% and 1.5\% better than PGA-Net (w/ CRFs) in ODS and AP metrics respectively, confirming 
the effectiveness of embedding an attention mechanism in the multi-scale CRF model. In Table~\ref{pascal-ablation}, the mIoU of PGA-Net (w/ FLAG-CRFs) is 1.39 points higher than that of PGA-Net (w/ CRFs), further demonstrating the advantage of the proposed attention mechanism. 
PGA-Net (w/o deep supervision) decreases the overall performance of our method by 1.9\% in ODS, showing the crucial importance of deep supervision
for better optimization of the whole PGA-Net. 
Comparing the performance of the proposed two versions of the AG-CRF model, \ie~PLAG-CRFs and FLAG-CRFs, we can see
that PGA-Net (FLAG-CRFs) slightly outperforms PGA-Net (PLAG-CRFs) in both ODS and OIS, while bringing a significant 
improvement (around 2\%) in AP. 

\begin{figure}[!t]
\centering
\includegraphics[width=0.93\linewidth]{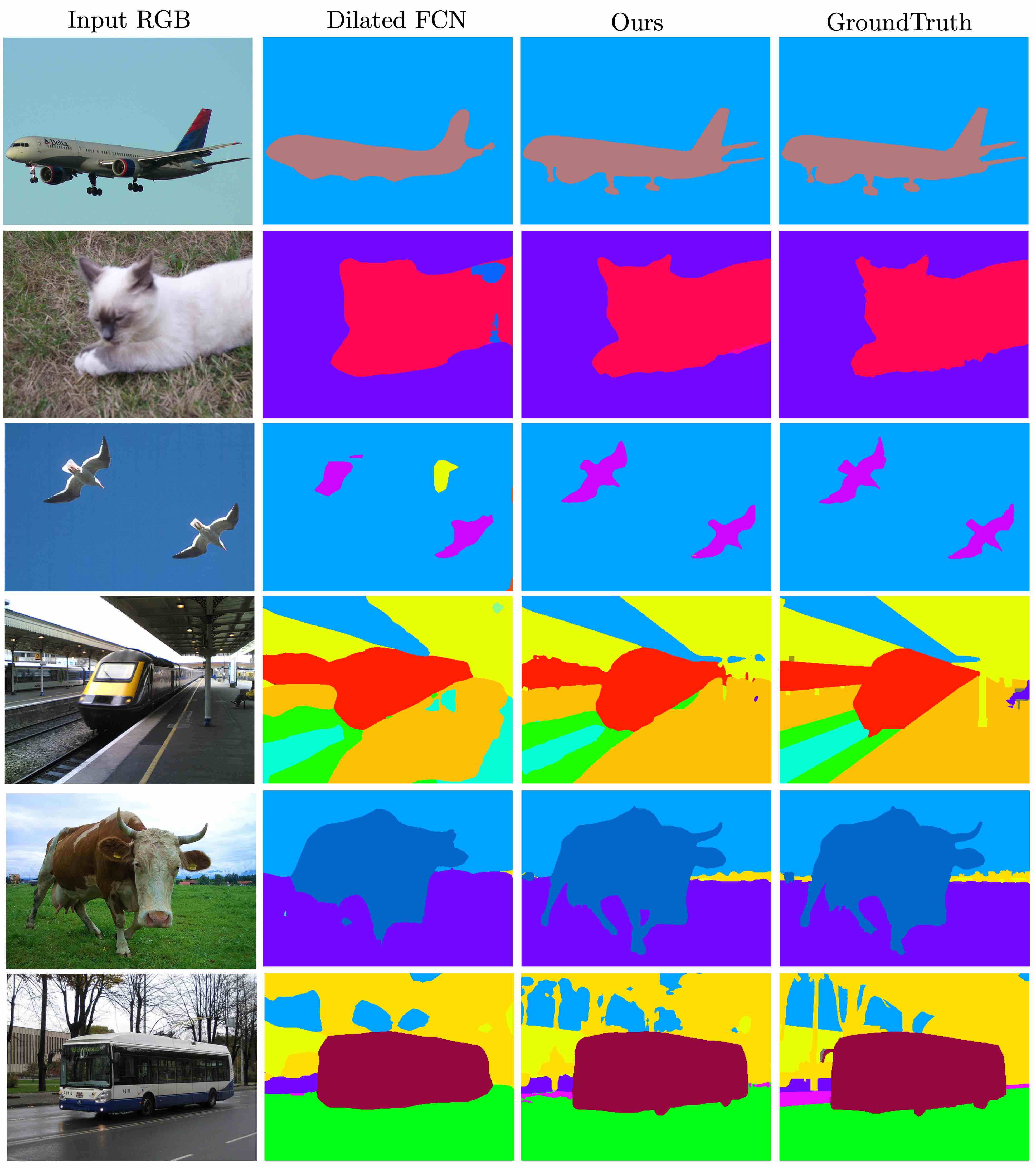}
\vspace{-5pt}
\caption{Qualitative semantic segmentation results on the Pascal-Context dataset. The representative Dilated FCN method~\cite{chen2016deeplab} is compared.}
\label{predvis:pcontext}
\vspace{-8pt}
\end{figure}

Finally, considering HED~\cite{xie2015holistically} and 
Hypercolumn~\cite{hariharan2015hypercolumns}, it is clear that our PGA-Net (FLAG-CRFs) is significantly better than these methods. 
Importantly, our approach utilizes only three scales while for HED~\cite{xie2015holistically} and 
Hypercolumn~\cite{hariharan2015hypercolumns} we consider five scales. \amid{By considering five scales, our model obtains further improvement upon the results using three scales for contour detection on the NYUD-V2 (see Table~\ref{nyud-ablation}) and for semantic segmenation on the Pascal-Context (see Table~\ref{pascal-ablation}). We also deploy the proposed AG-CRF model into a popular multi-scale network architecture HRNet~\cite{sun2019high}. It is clear to see that our model could also boost its performance, further confirming that our model is able to be applied into different multi-scale context for effective feature fusion.}
From Table~\ref{nyud-ablation} and~\ref{pascal-ablation}, we can also observe the effectiveness of the proposed conditional kernel strategy. PGA-Net (w/ FLAG-CRFs + CK) clearly improves over PGA-Net (w/ FLAG-CRFs) on the AP (1.3 points) for the contour detection and the mIoU (1.1 points) for the semantic segmentation. We also plot the training curves of the proposed approach and the baseline on the Pascal-Context validation in Figure~\ref{fig:comp_miou_pixelacc}. As shown in the figure, our approach consistently outperforms the baseline model at each training epoch, furthering demonstrating the effectiveness of the proposed AG-CRF model. 

\begin{table}
\centering
\caption{Overall performance comparison with state of the art methods on the \textit{val} set of the PASCAL-Context dataset. Spatial context post-processing is used in the pipeline of most methods except for~\cite{chen2016deeplab, xie2016top, zheng2015conditional}. Our full model achieves the best results compared with both the CRF-based or not the CRF-based approaches on the pixAcc and mIoU metrics. \amid{3S and 5S indicate three and five feature scales are considered, respectively.}}
\vspace{-5pt}
\resizebox{0.99\linewidth}{!} {
\begin{tabular}{l|c|c|cc}
\toprule[0.3ex]
Method & Backbone & pixAcc\% & mIoU\% \\\midrule
CFM (VGG+MCG)~\cite{dai2015convolutional}  & VGG-16  & - & 34.4  \\                                                         
FCN-8s~\cite{LongSD14}   & VGG-16 & 46.5 & 35.1  \\
FCN-8s~\cite{long2015fully}    & VGG-16 & 50.7 & 37.8 & -       \\
DeepLab-v2~\cite{chen2016deeplab}  & VGG-16 & - & 37.6   \\ 
BoxSup~\cite{dai2015boxsup}  & VGG-16 & - & 40.5   \\ 
ConvPP-8s~\cite{xie2016top} & VGG-16 & - & 41.0  \\
PixelNet~\cite{bansal2017pixelnet} & VGG-16 & 51.5 & 41.4   \\\midrule
CRF-RNN~\cite{zheng2015conditional} & VGG-16 & - & 39.3  \\
DeepLab-v2 + CRF~\cite{chen2016deeplab} & VGG-16 & -  & 45.7 \\ 
\midrule
ASPP~\cite{chen2017rethinking} & ResNet-50 & 78.3 & 49.2 \\
PSPNet~\cite{zhao2016pyramid} & ResNet-50 & 78.6 & 50.6 \\
EncNet~\cite{zhang2018context} & ResNet-50 & 78.4 & 49.9 \\
EncNet~\cite{zhang2018context} & ResNet-101 & 79.2 & 51.7 \\
\amid{HRNet~\cite{sun2019high}} & \amid{HRNetV2-W48} & \amid{-} & \amid{54.0} \\
\amid{OCR~\cite{yuan2019object}} & \amid{ResNet101} & \amid{-} & \amid{54.8} \\
\midrule
PGA-Net (w/ FLAG-CRFs+ CK) (3S) & ResNet-50 & {79.6} & {52.2} \\
PGA-Net (w/ FLAG-CRFs+ CK) (3S) & ResNet-101 & {80.8} & {54.8} \\
\amid{PGA-Net (w/ FLAG-CRFs+ CK) (5S)} & \amid{ResNet-101} & \amid{\textbf{81.2}} & \amid{\textbf{55.1}} \\
\bottomrule[0.3ex]                           
\end{tabular}
}
\label{overall_semantic}
\vspace{-2pt}
\end{table}

\begin{figure}[!t]
\centering
\includegraphics[width=0.9\linewidth]{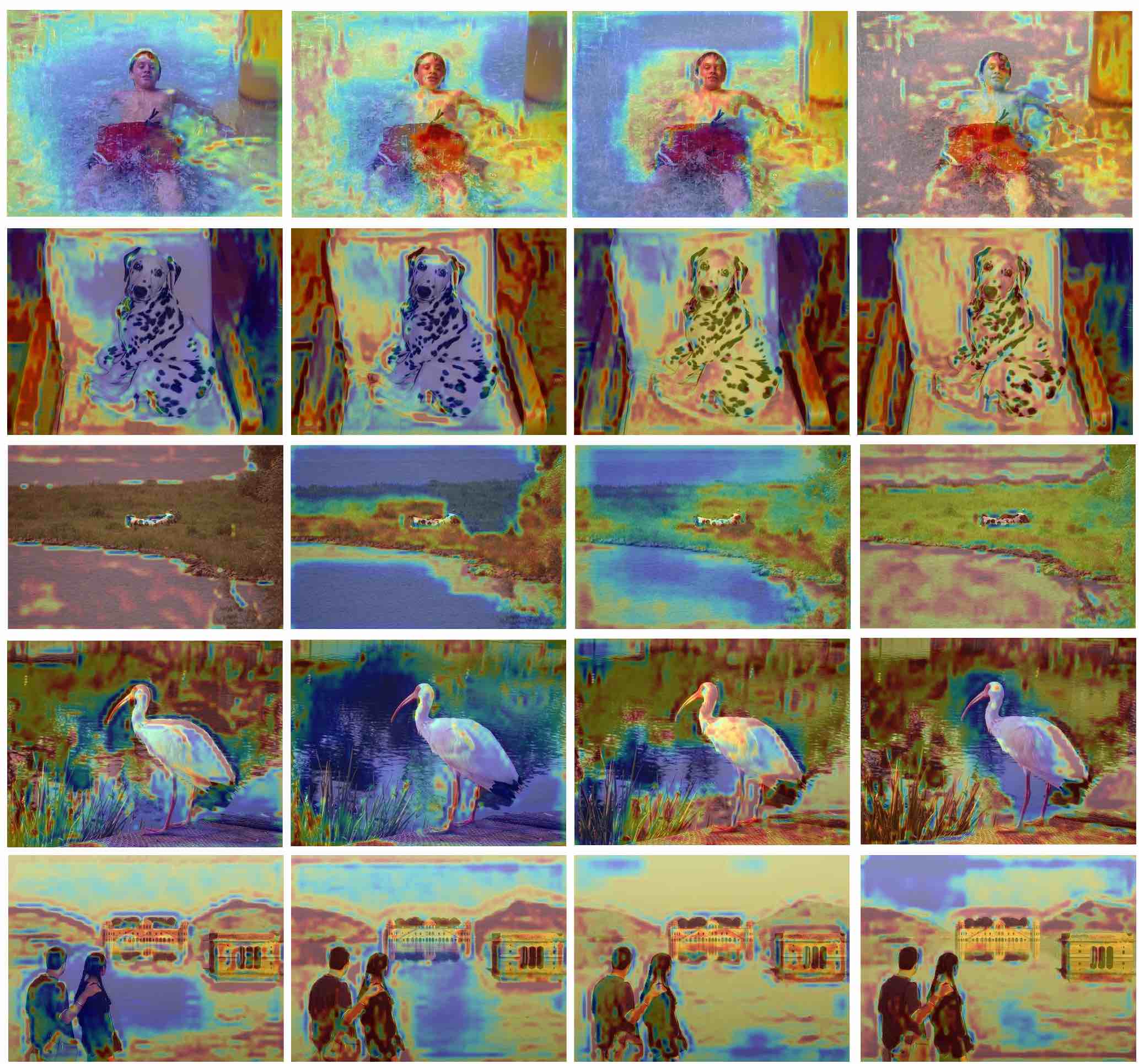}
\vspace{-5pt}
\caption{Visualization of the learned attention maps of the proposed AG-CRF model on the Pascal-Context dataset. Our model learns pixel-wise attention in both the spatial and channel dimension. We visualize the attention as we perform on KITTI by sampling four attention channels of the attention map uniformly. The learned attentions are able to capture different semantic regions to guide the message passing among features.}
\vspace{-8pt}
\label{fig:attention_pcontext}
\end{figure}

\subsubsection{Qualitative Analysis.} \par\noindent\textbf{Attention visualization.}
Figure~\ref{fig:attention_kitti} and Figure~\ref{fig:attention_pcontext} show examples of the learned attention maps in our proposed AG-CRF model on KITTI and Pascal-Context, respectively. As our attention mechanism learns a multi-channel attention map, meaning that the attention map has the same number of channels as the feature map. We visualize four channels of the overall 256 channels (\ie~every the 64-th channel). It can be observed that the learned attention map could capture the informative feature region from both the spatial and channel dimension, which we believe is an important reason for our model to effectively refine the feature maps. \amid{Taking the second row in Fig.~\ref{fig:attention_pcontext} for instance, it is easy to observe that the dog, the chair and the background are activated on different channels, and for each specific channel, different spatial regions are activated. The dark blue color marks the activated parts/regions.}
\par\noindent\textbf{Prediction visualization.} 
The multi-scale predictions and the final prediction from the PGA-Net on contour detection is shown in Figure~\ref{predvis:bsds}. It can be observed that the multi-scale predictions are highly complementary to each other, which confirming the initial intuition of modelling the multi-scale predictions in a joint CRF model for structured prediction and fusion. Figure~\ref{predvis:kitt} and Figure~\ref{predvis:pcontext} show examples of the monocular depth estimation on KITTI and the semantic segmentation on Pascal-Context respectively. Different state of the arts methods are compared in the figures. It is clearly that our approach achieves qualitatively better than these methods on both datasets. 

\section{Conclusion}\label{sec:conclusion}
We presented a novel 
multi-scale probabilistic graph attention networks with conditional kernels for pixel-wise prediction.
The proposed model introduces two main components, 
\ie~a hierarchical architecture for generating more rich and complementary multi-scale 
feature representations, and an Attention-Gated CRF model using conditional kernels for robust feature refinement and fusion. 
We demonstrate the effectiveness of our approach through extensive experiments on three different pixel-wise prediction tasks, including continuous problems, \ie~monocular depth estimation, and discrete problems, \ie~object contour detection and semantic segmentation. Four challenging publicly available datasets, BSDS500, NYUD-V2, KITTI and Pascal-Context are considered in our experiments. The proposed model achieved very competitive performance on all the task and the datasets.  
The proposed approach addresses a general problem, \ie~how to learn rich multi-scale representations and 
optimally fuse them. Therefore, we believe it may be also beneficial for other continuous and discrete pixel-level prediction tasks.



\ifCLASSOPTIONcaptionsoff
  \newpage
\fi



%
%
%

\small
\bibliographystyle{IEEEtran}
\bibliography{nips_pami} 

\begin{thebibliography}{100}
\providecommand{\url}[1]{#1}
\def\UrlFont{\rmfamily}
\providecommand{\newblock}{\relax}
\providecommand{\bibinfo}[2]{#2}
\providecommand\BIBentrySTDinterwordspacing{\spaceskip=0pt\relax}
\providecommand\BIBentryALTinterwordstretchfactor{4}
\providecommand\BIBentryALTinterwordspacing{\spaceskip=\fontdimen2\font plus
\BIBentryALTinterwordstretchfactor\fontdimen3\font minus
  \fontdimen4\font\relax}
\providecommand\BIBforeignlanguage[2]{{%
\expandafter\ifx\csname l@#1\endcsname\relax
\typeout{** WARNING: IEEEtran.bst: No hyphenation pattern has been}%
\typeout{** loaded for the language `#1'. Using the pattern for}%
\typeout{** the default language instead.}%
\else
\language=\csname l@#1\endcsname
\fi
#2}}

\bibitem{ren2008multi}
X.~Ren, ``Multi-scale improves boundary detection in natural images,'' in
  \emph{ECCV}, 2008.

\bibitem{xie2015holistically}
S.~Xie and Z.~Tu, ``Holistically-nested edge detection,'' in \emph{CVPR}, 2015.

\bibitem{kokkinos2015pushing}
I.~Kokkinos, ``Pushing the boundaries of boundary detection using deep
  learning,'' in \emph{ICLR}, 2016.

\bibitem{maninis2016convolutional}
K.-K. Maninis, J.~Pont-Tuset, P.~Arbel{\'a}ez, and L.~Van~Gool, ``Convolutional
  oriented boundaries,'' in \emph{ECCV}, 2016.

\bibitem{chen2016deeplab}
L.-C. Chen, G.~Papandreou, I.~Kokkinos, K.~Murphy, and A.~L. Yuille, ``Deeplab:
  Semantic image segmentation with deep convolutional nets, atrous convolution,
  and fully connected crfs,'' \emph{arXiv preprint arXiv:1606.00915}, 2016.

\bibitem{liu2015deep}
F.~Liu, C.~Shen, and G.~Lin, ``Deep convolutional neural fields for depth
  estimation from a single image,'' in \emph{CVPR}, 2015.

\bibitem{xu2017multi}
D.~Xu, E.~Ricci, W.~Ouyang, X.~Wang, and N.~Sebe, ``Multi-scale continuous crfs
  as sequential deep networks for monocular depth estimation,'' in \emph{CVPR},
  2017.

\bibitem{mnih2014recurrent}
V.~Mnih, N.~Heess, A.~Graves, \emph{et~al.}, ``Recurrent models of visual
  attention,'' in \emph{NIPS}, 2014.

\bibitem{minka2009gates}
T.~Minka and J.~Winn, ``Gates,'' in \emph{NIPS}, 2009.

\bibitem{arbelaez2011contour}
P.~Arbel{\'a}ez, M.~Maire, C.~Fowlkes, and J.~Malik, ``Contour detection and
  hierarchical image segmentation,'' \emph{TPAMI}, vol.~33, no.~5, pp.
  898--916, 2011.

\bibitem{silberman2012indoor}
N.~Silberman, D.~Hoiem, P.~Kohli, and R.~Fergus, ``Indoor segmentation and
  support inference from rgbd images,'' in \emph{ECCV}, 2012.

\bibitem{Geiger2013IJRR}
A.~Geiger, P.~Lenz, C.~Stiller, and R.~Urtasun, ``Vision meets robotics: The
  kitti dataset,'' \emph{IJRR}, vol.~32, no.~11, pp. 1231--1237, 2013.

\bibitem{mottaghi2014role}
R.~Mottaghi, X.~Chen, X.~Liu, N.-G. Cho, S.-W. Lee, S.~Fidler, R.~Urtasun, and
  A.~Yuille, ``The role of context for object detection and semantic
  segmentation in the wild,'' in \emph{CVPR}, 2014.

\bibitem{xu2017learningdeep}
D.~Xu, W.~Ouyang, X.~Alameda-Pineda, E.~Ricci, X.~Wang, and N.~Sebe, ``Learning
  deep structured multi-scale features using attention-gated crfs for contour
  prediction,'' in \emph{NIPS}, 2017.

\bibitem{hariharan2015hypercolumns}
B.~Hariharan, P.~Arbel{\'a}ez, R.~Girshick, and J.~Malik, ``Hypercolumns for
  object segmentation and fine-grained localization,'' in \emph{CVPR}, 2015.

\bibitem{shen2015deepcontour}
W.~Shen, X.~Wang, Y.~Wang, X.~Bai, and Z.~Zhang, ``Deepcontour: A deep
  convolutional feature learned by positive-sharing loss for contour
  detection,'' in \emph{CVPR}, 2015.

\bibitem{bertasius2015deepedge}
G.~Bertasius, J.~Shi, and L.~Torresani, ``Deepedge: A multi-scale bifurcated
  deep network for top-down contour detection,'' in \emph{CVPR}, 2015.

\bibitem{yang2016object}
J.~Yang, B.~Price, S.~Cohen, H.~Lee, and M.-H. Yang, ``Object contour detection
  with a fully convolutional encoder-decoder network,'' in \emph{CVPR}, 2016.

\bibitem{liu2016richer}
Y.~Liu, M.-M. Cheng, X.~Hu, K.~Wang, and X.~Bai, ``Richer convolutional
  features for edge detection,'' \emph{arXiv preprint arXiv:1612.02103}, 2016.

\bibitem{wang2018deep}
Y.~Wang, X.~Zhao, Y.~Li, and K.~Huang, ``Deep crisp boundaries: From boundaries
  to higher-level tasks,'' \emph{TIP}, vol.~28, no.~3, pp. 1285--1298, 2018.

\bibitem{eigen2015predicting}
D.~Eigen and R.~Fergus, ``Predicting depth, surface normals and semantic labels
  with a common multi-scale convolutional architecture,'' in \emph{ICCV}, 2015.

\bibitem{wang2015towards}
P.~Wang, X.~Shen, Z.~Lin, S.~Cohen, B.~Price, and A.~Yuille, ``Towards unified
  depth and semantic prediction from a single image,'' in \emph{CVPR}, 2015.

\bibitem{roymonocular}
A.~Roy and S.~Todorovic, ``Monocular depth estimation using neural regression
  forest,'' in \emph{CVPR}, 2016.

\bibitem{laina2016deeper}
I.~Laina, C.~Rupprecht, V.~Belagiannis, F.~Tombari, and N.~Navab, ``Deeper
  depth prediction with fully convolutional residual networks,'' \emph{arXiv
  preprint arXiv:1606.00373}, 2016.

\bibitem{fu2018deep}
H.~Fu, M.~Gong, C.~Wang, K.~Batmanghelich, and D.~Tao, ``Deep ordinal
  regression network for monocular depth estimation,'' in \emph{CVPR}, 2018.

\bibitem{gan2018monocular}
Y.~Gan, X.~Xu, W.~Sun, and L.~Lin, ``Monocular depth estimation with affinity,
  vertical pooling, and label enhancement,'' in \emph{ECCV}, 2018.

\bibitem{lee2019big}
J.~H. Lee, M.-K. Han, D.~W. Ko, and I.~H. Suh, ``From big to small: Multi-scale
  local planar guidance for monocular depth estimation,'' \emph{arXiv preprint
  arXiv:1907.10326}, 2019.

\bibitem{eigen2014depth}
D.~Eigen, C.~Puhrsch, and R.~Fergus, ``Depth map prediction from a single image
  using a multi-scale deep network,'' in \emph{NIPS}, 2014.

\bibitem{long2015fully}
J.~Long, E.~Shelhamer, and T.~Darrell, ``Fully convolutional networks for
  semantic segmentation,'' in \emph{CVPR}, 2015.

\bibitem{yu2015multi}
F.~Yu and V.~Koltun, ``Multi-scale context aggregation by dilated
  convolutions,'' \emph{arXiv preprint arXiv:1511.07122}, 2015.

\bibitem{yuan2018ocnet}
Y.~Yuan and J.~Wang, ``Ocnet: Object context network for scene parsing,''
  \emph{arXiv preprint arXiv:1809.00916}, 2018.

\bibitem{chen2016attention}
L.-C. Chen, Y.~Yang, J.~Wang, W.~Xu, and A.~L. Yuille, ``Attention to scale:
  Scale-aware semantic image segmentation,'' in \emph{CVPR}, 2016.

\bibitem{noh2015learning}
H.~Noh, S.~Hong, and B.~Han, ``Learning deconvolution network for semantic
  segmentation,'' in \emph{ICCV}, 2015.

\bibitem{badrinarayanan2015segnet}
V.~Badrinarayanan, A.~Handa, and R.~Cipolla, ``Segnet: A deep convolutional
  encoder-decoder architecture for robust semantic pixel-wise labelling,''
  \emph{arXiv preprint arXiv:1505.07293}, 2015.

\bibitem{liu2015semantic}
Z.~Liu, X.~Li, P.~Luo, C.-C. Loy, and X.~Tang, ``Semantic image segmentation
  via deep parsing network,'' in \emph{ICCV}, 2015.

\bibitem{arnab2016higher}
A.~Arnab, S.~Jayasumana, S.~Zheng, and P.~H. Torr, ``Higher order conditional
  random fields in deep neural networks,'' in \emph{ECCV}, 2016.

\bibitem{zheng2015conditional}
S.~Zheng, S.~Jayasumana, B.~Romera-Paredes, V.~Vineet, Z.~Su, D.~Du, C.~Huang,
  and P.~H. Torr, ``Conditional random fields as recurrent neural networks,''
  in \emph{ICCV}, 2015.

\bibitem{chen2019graph}
Y.~Chen, M.~Rohrbach, Z.~Yan, Y.~Shuicheng, J.~Feng, and Y.~Kalantidis,
  ``Graph-based global reasoning networks,'' in \emph{CVPR}, 2019.

\bibitem{xu2018pad}
D.~Xu, W.~Ouyang, X.~Wang, and N.~Sebe, ``Pad-net: Multi-tasks guided
  prediction-and-distillation network for simultaneous depth estimation and
  scene parsing,'' in \emph{CVPR}, 2018.

\bibitem{zhang2018joint}
Z.~Zhang, Z.~Cui, C.~Xu, Z.~Jie, X.~Li, and J.~Yang, ``Joint task-recursive
  learning for semantic segmentation and depth estimation,'' in \emph{ECCV},
  2018.

\bibitem{zhang2019pattern}
Z.~Zhang, Z.~Cui, C.~Xu, Y.~Yan, N.~Sebe, and J.~Yang, ``Pattern-affinitive
  propagation across depth, surface normal and semantic segmentation,'' in
  \emph{CVPR}, 2019.

\bibitem{vandenhende2020mti}
S.~Vandenhende, S.~Georgoulis, and L.~Van~Gool, ``Mti-net: Multi-scale task
  interaction networks for multi-task learning,'' in \emph{ECCV}, 2020.

\bibitem{sun2019high}
K.~Sun, Y.~Zhao, B.~Jiang, T.~Cheng, B.~Xiao, D.~Liu, Y.~Mu, X.~Wang, W.~Liu,
  and J.~Wang, ``High-resolution representations for labeling pixels and
  regions,'' \emph{arXiv preprint arXiv:1904.04514}, 2019.

\bibitem{wang2020deep}
J.~Wang, K.~Sun, T.~Cheng, B.~Jiang, C.~Deng, Y.~Zhao, D.~Liu, Y.~Mu, M.~Tan,
  X.~Wang, \emph{et~al.}, ``Deep high-resolution representation learning for
  visual recognition,'' \emph{TPAMI}, 2020.

\bibitem{chen2014semantic}
L.-C. Chen, G.~Papandreou, I.~Kokkinos, K.~Murphy, and A.~L. Yuille, ``Semantic
  image segmentation with deep convolutional nets and fully connected crfs,''
  in \emph{ICLR}, 2015.

\bibitem{yang2015multi}
S.~Yang and D.~Ramanan, ``Multi-scale recognition with dag-cnns,'' in
  \emph{ICCV}, 2015.

\bibitem{huang2018multi}
G.~Huang and D.~Chen, ``Multi-scale dense networks for resource efficient image
  classification,'' in \emph{ICLR}, 2018.

\bibitem{velivckovic2017graph}
P.~Veli{\v{c}}kovi{\'c}, G.~Cucurull, A.~Casanova, A.~Romero, P.~Lio, and
  Y.~Bengio, ``Graph attention networks,'' in \emph{ICLR}, 2017.

\bibitem{xiao2015application}
T.~Xiao, Y.~Xu, K.~Yang, J.~Zhang, Y.~Peng, and Z.~Zhang, ``The application of
  two-level attention models in deep convolutional neural network for
  fine-grained image classification,'' in \emph{CVPR}, 2015.

\bibitem{chorowski2015attention}
J.~K. Chorowski, D.~Bahdanau, D.~Serdyuk, K.~Cho, and Y.~Bengio,
  ``Attention-based models for speech recognition,'' in \emph{NIPS}, 2015.

\bibitem{xu2015show}
K.~Xu, J.~Ba, R.~Kiros, K.~Cho, A.~Courville, R.~Salakhudinov, R.~Zemel, and
  Y.~Bengio, ``Show, attend and tell: Neural image caption generation with
  visual attention,'' in \emph{ICML}, 2015.

\bibitem{vaswani2017attention}
A.~Vaswani, N.~Shazeer, N.~Parmar, J.~Uszkoreit, L.~Jones, A.~N. Gomez,
  {\L}.~Kaiser, and I.~Polosukhin, ``Attention is all you need,'' in
  \emph{NIPS}, 2017.

\bibitem{fu2019dual}
J.~Fu, J.~Liu, H.~Tian, Y.~Li, Y.~Bao, Z.~Fang, and H.~Lu, ``Dual attention
  network for scene segmentation,'' in \emph{CVPR}, 2019.

\bibitem{tang2010gated}
Y.~Tang, ``Gated boltzmann machine for recognition under occlusion,'' in
  \emph{NIPS Workshop on Transfer Learning by Learning Rich Generative Models},
  2010.

\bibitem{chung2014empirical}
J.~Chung, C.~Gulcehre, K.~Cho, and Y.~Bengio, ``Empirical evaluation of gated
  recurrent neural networks on sequence modeling,'' \emph{arXiv preprint
  arXiv:1412.3555}, 2014.

\bibitem{zeng2016crafting}
X.~Zeng, W.~Ouyang, J.~Yan, H.~Li, T.~Xiao, K.~Wang, Y.~Liu, Y.~Zhou, B.~Yang,
  Z.~Wang, \emph{et~al.}, ``Crafting gbd-net for object detection,''
  \emph{arXiv preprint arXiv:1610.02579}, 2016.

\bibitem{quattoni2005conditional}
A.~Quattoni, M.~Collins, and T.~Darrell, ``Conditional random fields for object
  recognition,'' in \emph{NIPS}, 2005.

\bibitem{sarawagi2005semi}
S.~Sarawagi and W.~W. Cohen, ``Semi-markov conditional random fields for
  information extraction,'' in \emph{NIPS}, 2005.

\bibitem{lafferty2001conditional}
J.~Lafferty, A.~McCallum, and F.~C. Pereira, ``Conditional random fields:
  Probabilistic models for segmenting and labeling sequence data,'' in
  \emph{ICML}, 2001.

\bibitem{he2004multiscale}
X.~He, R.~S. Zemel, and M.~{\'A}. Carreira-Perpi{\~n}{\'a}n, ``Multiscale
  conditional random fields for image labeling,'' in \emph{CVPR}, 2004.

\bibitem{krahenbuhl2011efficient}
P.~Kr{\"a}henb{\"u}hl and V.~Koltun, ``Efficient inference in fully connected
  crfs with gaussian edge potentials,'' in \emph{NIPS}, 2011.

\bibitem{boykov2006graph}
Y.~Boykov and G.~Funka-Lea, ``Graph cuts and efficient nd image segmentation,''
  \emph{IJCV}, vol.~70, no.~2, pp. 109--131, 2006.

\bibitem{chu2016crf}
X.~Chu, W.~Ouyang, X.~Wang, \emph{et~al.}, ``Crf-cnn: Modeling structured
  information in human pose estimation,'' in \emph{NIPS}, 2016.

\bibitem{winn2012causality}
J.~Winn, ``Causality with gates,'' in \emph{AISTATS}, 2012.

\bibitem{krizhevsky2012imagenet}
A.~Krizhevsky, I.~Sutskever, and G.~E. Hinton, ``Imagenet classification with
  deep convolutional neural networks,'' in \emph{NIPS}, 2012.

\bibitem{simonyan2014very}
K.~Simonyan and A.~Zisserman, ``Very deep convolutional networks for
  large-scale image recognition,'' \emph{arXiv preprint arXiv:1409.1556}, 2014.

\bibitem{He2015}
K.~He, X.~Zhang, S.~Ren, and J.~Sun, ``Deep residual learning for image
  recognition,'' \emph{arXiv preprint arXiv:1512.03385}, 2015.

\bibitem{zhang2018context}
H.~Zhang, K.~Dana, J.~Shi, Z.~Zhang, X.~Wang, A.~Tyagi, and A.~Agrawal,
  ``Context encoding for semantic segmentation,'' in \emph{CVPR}, 2018.

\bibitem{garg2016unsupervised}
R.~Garg, G.~Carneiro, and I.~Reid, ``Unsupervised cnn for single view depth
  estimation: Geometry to the rescue,'' in \emph{ECCV}, 2016.

\bibitem{felzenszwalb2004efficient}
P.~F. Felzenszwalb and D.~P. Huttenlocher, ``Efficient graph-based image
  segmentation,'' \emph{IJCV}, vol.~59, no.~2, 2004.

\bibitem{comaniciu2002mean}
D.~Comaniciu and P.~Meer, ``Mean shift: A robust approach toward feature space
  analysis,'' \emph{TPAMI}, vol.~24, no.~5, pp. 603--619, 2002.

\bibitem{shi2000normalized}
J.~Shi and J.~Malik, ``Normalized cuts and image segmentation,'' \emph{TPAMI},
  vol.~22, no.~8, 2000.

\bibitem{ren2013image}
Z.~Ren and G.~Shakhnarovich, ``Image segmentation by cascaded region
  agglomeration,'' in \emph{CVPR}, 2013.

\bibitem{lim2013sketch}
J.~J. Lim, C.~L. Zitnick, and P.~Doll{\'a}r, ``Sketch tokens: A learned
  mid-level representation for contour and object detection,'' in \emph{CVPR},
  2013.

\bibitem{pont2015multiscale}
J.~Pont-Tuset, P.~Arbelaez, J.~T. Barron, F.~Marques, and J.~Malik,
  ``Multiscale combinatorial grouping for image segmentation and object
  proposal generation,'' \emph{TPAMI}, vol.~39, no.~1, pp. 128--140, 2016.

\bibitem{zhaosegmenting}
Q.~Zhao, ``Segmenting natural images with the least effort as humans,'' in
  \emph{BMVC}, 2015.

\bibitem{dollar2013structured}
P.~Doll{\'a}r and C.~L. Zitnick, ``Structured forests for fast edge
  detection,'' in \emph{ICCV}, 2013.

\bibitem{gupta2014learning}
S.~Gupta, R.~Girshick, P.~Arbel{\'a}ez, and J.~Malik, ``Learning rich features
  from rgb-d images for object detection and segmentation,'' in \emph{ECCV},
  2014.

\bibitem{zhou2017scene}
B.~Zhou, H.~Zhao, X.~Puig, S.~Fidler, A.~Barriuso, and A.~Torralba, ``Scene
  parsing through ade20k dataset,'' in \emph{CVPR}, 2017.

\bibitem{deng2009imagenet}
J.~Deng, W.~Dong, R.~Socher, L.-J. Li, K.~Li, and L.~Fei-Fei, ``Imagenet: A
  large-scale hierarchical image database,'' in \emph{CVPR}, 2009.

\bibitem{godard2016unsupervised}
C.~Godard, O.~Mac~Aodha, and G.~J. Brostow, ``Unsupervised monocular depth
  estimation with left-right consistency,'' in \emph{CVPR}, 2017.

\bibitem{xu2018monocular}
D.~Xu, E.~Ricci, W.~Ouyang, X.~Wang, N.~Sebe, \emph{et~al.}, ``Monocular depth
  estimation using multi-scale continuous crfs as sequential deep networks,''
  \emph{TPAMI}, vol.~41, no.~6, pp. 1426--1440, 2018.

\bibitem{hallman2015oriented}
S.~Hallman and C.~C. Fowlkes, ``Oriented edge forests for boundary detection,''
  in \emph{CVPR}, 2015.

\bibitem{zhang2016semicontour}
Z.~Zhang, F.~Xing, X.~Shi, and L.~Yang, ``Semicontour: A semi-supervised
  learning approach for contour detection,'' in \emph{CVPR}, 2016.

\bibitem{dollar2015fast}
P.~Doll{\'a}r and C.~L. Zitnick, ``Fast edge detection using structured
  forests,'' \emph{TPAMI}, vol.~37, no.~8, pp. 1558--1570, 2015.

\bibitem{gupta2013perceptual}
S.~Gupta, P.~Arbelaez, and J.~Malik, ``Perceptual organization and recognition
  of indoor scenes from rgb-d images,'' in \emph{CVPR}, 2013.

\bibitem{zhou2017unsupervised}
T.~Zhou, M.~Brown, N.~Snavely, and D.~G. Lowe, ``Unsupervised learning of depth
  and ego-motion from video,'' in \emph{CVPR}, 2017.

\bibitem{adadepth}
J.~N. Kundu, P.~K. Uppala, A.~Pahuja, and R.~V. Babu, ``Adadepth: Unsupervised
  content congruent adaptation for depth estimation,'' in \emph{CVPR}, 2018.

\bibitem{pilzer2018unsupervised}
A.~Pilzer, D.~Xu, M.~Puscas, E.~Ricci, and N.~Sebe, ``Unsupervised adversarial
  depth estimation using cycled generative networks,'' in \emph{3DV}, 2018.

\bibitem{wang2017learning}
C.~Wang, J.~M. Buenaposada, R.~Zhu, and S.~Lucey, ``Learning depth from
  monocular videos using direct methods,'' in \emph{CVPR}, 2018.

\bibitem{zou2018df}
Y.~Zou, Z.~Luo, and J.-B. Huang, ``Df-net: Unsupervised joint learning of depth
  and flow using cross-task consistency,'' in \emph{ECCV}, 2018.

\bibitem{zhan2018unsupervised}
H.~Zhan, R.~Garg, C.~S. Weerasekera, K.~Li, H.~Agarwal, and I.~Reid,
  ``Unsupervised learning of monocular depth estimation and visual odometry
  with deep feature reconstruction,'' \emph{arXiv preprint arXiv:1803.03893},
  2018.

\bibitem{kuznietsov2017semi}
Y.~Kuznietsov, J.~St{\"u}ckler, and B.~Leibe, ``Semi-supervised deep learning
  for monocular depth map prediction,'' in \emph{CVPR}, 2017.

\bibitem{saxena2009make3d}
A.~Saxena, M.~Sun, and A.~Y. Ng, ``Make3d: Learning 3d scene structure from a
  single still image,'' \emph{TPAMI}, vol.~31, no.~5, pp. 824--840, 2009.

\bibitem{mahjourian2018unsupervised}
R.~Mahjourian, M.~Wicke, and A.~Angelova, ``Unsupervised learning of depth and
  ego-motion from monocular video using 3d geometric constraints,'' in
  \emph{CVPR}, 2018.

\bibitem{chen2017rethinking}
L.-C. Chen, G.~Papandreou, F.~Schroff, and H.~Adam, ``Rethinking atrous
  convolution for semantic image segmentation,'' \emph{arXiv preprint
  arXiv:1706.05587}, 2017.

\bibitem{zhao2016pyramid}
H.~Zhao, J.~Shi, X.~Qi, X.~Wang, and J.~Jia, ``Pyramid scene parsing network,''
  \emph{arXiv preprint arXiv:1612.01105}, 2016.

\bibitem{xie2016top}
S.~Xie, X.~Huang, and Z.~Tu, ``Top-down learning for structured labeling with
  convolutional pseudoprior,'' in \emph{ECCV}, 2016.

\bibitem{dai2015convolutional}
J.~Dai, K.~He, and J.~Sun, ``Convolutional feature masking for joint object and
  stuff segmentation,'' in \emph{CVPR}, 2015.

\bibitem{LongSD14}
J.~Long, E.~Shelhamer, and T.~Darrell, ``Fully convolutional networks for
  semantic segmentation,'' in \emph{CVPR}, 2015.

\bibitem{dai2015boxsup}
J.~Dai, K.~He, and J.~Sun, ``Boxsup: Exploiting bounding boxes to supervise
  convolutional networks for semantic segmentation,'' in \emph{ICCV}, 2015.

\bibitem{bansal2017pixelnet}
A.~Bansal, X.~Chen, B.~Russell, A.~Gupta, and D.~Ramanan, ``Pixelnet:
  Representation of the pixels, by the pixels, and for the pixels,''
  \emph{arXiv preprint arXiv:1702.06506}, 2017.

\bibitem{yuan2019object}
Y.~Yuan, X.~Chen, and J.~Wang, ``Object-contextual representations for semantic
  segmentation,'' \emph{arXiv preprint arXiv:1909.11065}, 2019.

\end{thebibliography}

%
\vspace{-26pt}
\begin{IEEEbiography}[{\includegraphics[width=1in,height=1.25in,clip,keepaspectratio]{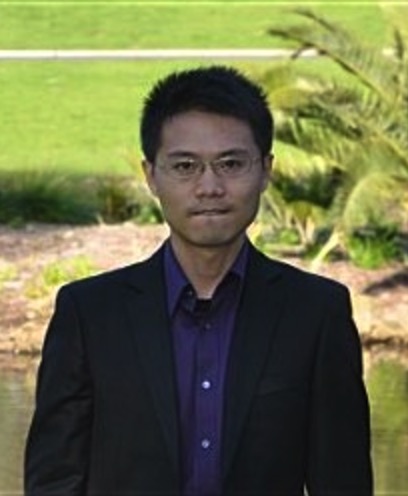}}]
{Dan Xu}
is an Assistant Professor in the Department of Computer Science and Engineering at HKUST. He was a Postdoctoral Research Fellow in VGG at the University of Oxford. He was a Ph.D. in the Department of Computer Science at the University of Trento. He was also a research assistant of MM Lab at the Chinese University of Hong Kong. He received the best scientific paper award at ICPR 2016, and a Best Paper Nominee at ACM MM 2018. He served as Area Chairs of ACM MM 2020, ICPR 2020 and WACV 2021.
\end{IEEEbiography}
\vspace{-20pt}
\begin{IEEEbiography}[{\includegraphics[width=1in,height=1.25in,clip,keepaspectratio]{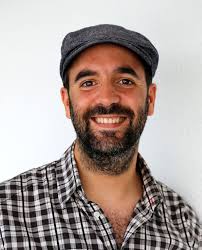}}]
{Xavier Alameda-Pineda} received M.Sc. degrees in mathematics (2008), in
telecommunications (2009) and in computer science
(2010) and a Ph.D. in mathematics and computer
science (2013) from Université Joseph Fourier. Since
2016, he is a Research Scientist at Inria Grenoble
Rhône-Alpes, with the Perception team. He served as
Area Chair at ICCV’17, of ICIAP’19 and of ACM
MM’19. He is the recipient of several paper awards
and of the ACM SIGMM Rising Star Award in 2018.
\end{IEEEbiography}
\vspace{-20pt}
\begin{IEEEbiography}[{\includegraphics[width=1in,height=1.25in,clip,keepaspectratio]{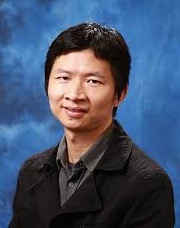}}]
{Wanli Ouyang}
received the PhD degree in the
Department of Electronic Engineering, The Chinese
University of Hong Kong. He is now
a senior lecturer in the School of Electrical and Information Engineering at the University of Sydney, Australia. His research interests include image processing, computer vision and pattern recognition. He is a senior member of IEEE.
\end{IEEEbiography}
\vspace{-20pt}
\begin{IEEEbiography}[{\includegraphics[width=1in,height=1.25in,clip,keepaspectratio]{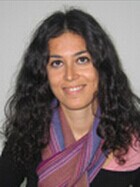}}]
{Elisa Ricci}
received the PhD degree from the
University of Perugia in 2008. She is an associate
professor at the University of Trento and a
researcher at Fondazione Bruno Kessler. She
has since been a post-doctoral researcher at
Idiap, Martigny, and Fondazione Bruno Kessler,
Trento. She was also a visiting researcher at the
University of Bristol. Her research interests are
mainly in the areas of computer vision and
machine learning. She is a member of the IEEE.
\end{IEEEbiography}
\vspace{-20pt}
\begin{IEEEbiography}[{\includegraphics[width=1in,height=1.25in,clip,keepaspectratio]{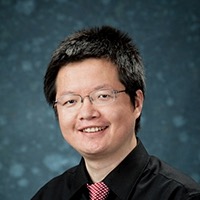}}]
{Xiaogang Wang}
received the PhD degree in Computer Science from Massachusetts Institute of Technology. He is an associate professor in the Department of Electronic Engineering at the Chinese University of Hong Kong since August 2009. He was the Area Chairs of ICCV 2011 and 2015, ECCV 2014 and 2016, ACCV 2014 and 2016. He received the Outstanding Young Researcher in Automatic Human Behaviour Analysis Award in 2011, Hong Kong RGC Early Career Award in 2012, and CUHK Young Researcher Award 2012.
\end{IEEEbiography}
\vspace{-20pt}

\begin{IEEEbiography}[{\includegraphics[width=1in,height=1.25in,clip,keepaspectratio]{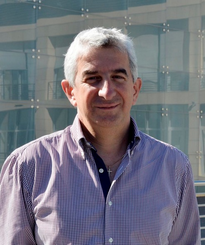}}]
{Nicu Sebe} is Professor with the University of Trento, Italy, leading the research in the areas of multimedia information retrieval and human behavior understanding. He was the General Co- Chair of the IEEE FG Conference 2008 and ACM Multimedia 2013, and the Program Chair of the International Conference on Image and Video Retrieval in 2007 and 2010, ACM Multimedia 2007 and 2011. He was the Program Chair of ICCV 2017 and ECCV 2016,
and a General Chair of ACM ICMR 2017. He is a fellow of the IAPR.
\end{IEEEbiography}




\end{document}